\documentclass[conference]{IEEEtran}
\IEEEoverridecommandlockouts
\usepackage{cite}
\usepackage{amsmath,amssymb,amsfonts}
\usepackage{algorithmic}
\usepackage{graphicx}
\usepackage{tabularx}
\usepackage{textcomp}
\usepackage{xcolor}
\usepackage{subcaption}
\usepackage{hyperref}
\def\BibTeX{{\rm B\kern-.05em{\sc i\kern-.025em b}\kern-.08em
    T\kern-.1667em\lower.7ex\hbox{E}\kern-.125emX}}
\begin{document}

\newcommand\blfootnote[1]{%
  \begingroup
  \renewcommand\thefootnote{}\footnote{#1}%
  \addtocounter{footnote}{-1}%
  \endgroup
}

\title{Characterizing and Taming Resolution in Convolutional Neural Networks\\
}

\author{\IEEEauthorblockN{Eddie Yan}
\IEEEauthorblockA{\textit{Allen School} \\
\textit{University of Washington}\\
Seattle, USA \\
eqy@cs.washington.edu}
\and
\IEEEauthorblockN{Liang Luo}
\IEEEauthorblockA{\textit{Allen School} \\
\textit{University of Washington}\\
Seattle, USA \\
liangluo@cs.washington.edu}
\and
\IEEEauthorblockN{Luis Ceze}
\IEEEauthorblockA{\textit{Allen School} \\
\textit{University of Washington}\\
Seattle, USA \\
luisceze@cs.washington.edu}
}

\maketitle
\thispagestyle{plain}
\pagestyle{plain}

\begin{abstract}
Image resolution has a significant effect on the accuracy and computational, storage, and bandwidth costs of computer vision model inference.
These costs are exacerbated when scaling out models to large inference serving systems and make image resolution an attractive target for optimization.
However, the choice of resolution inherently introduces additional tightly coupled choices, such as image crop size, image detail, and compute kernel implementation that impact computational, storage, and bandwidth costs.
Further complicating this setting, the optimal choices from the perspective of these metrics are highly dependent on the dataset and problem scenario.
We characterize this tradeoff space, quantitatively studying the accuracy and efficiency tradeoff via  systematic and automated tuning of image resolution, image quality and convolutional neural network operators. With the insights from this study, we propose a dynamic resolution mechanism that removes the need to statically choose a resolution ahead of time. 
Our evaluation shows that our dynamic resolution approach improves inference latency by $1.2\times$$-$$1.7\times$, reduces data access volume by up to 20--30\%, without affecting accuracy. We establish the dynamic resolution approach as a viable alternative to fine-tuning for a specific object scale to compensate for unknown crop sizes, which is the current state of the art. 
\end{abstract}

\section{Introduction}
 \begin{figure}[!ht]
    \centering
    \includegraphics[width=0.45\textwidth]{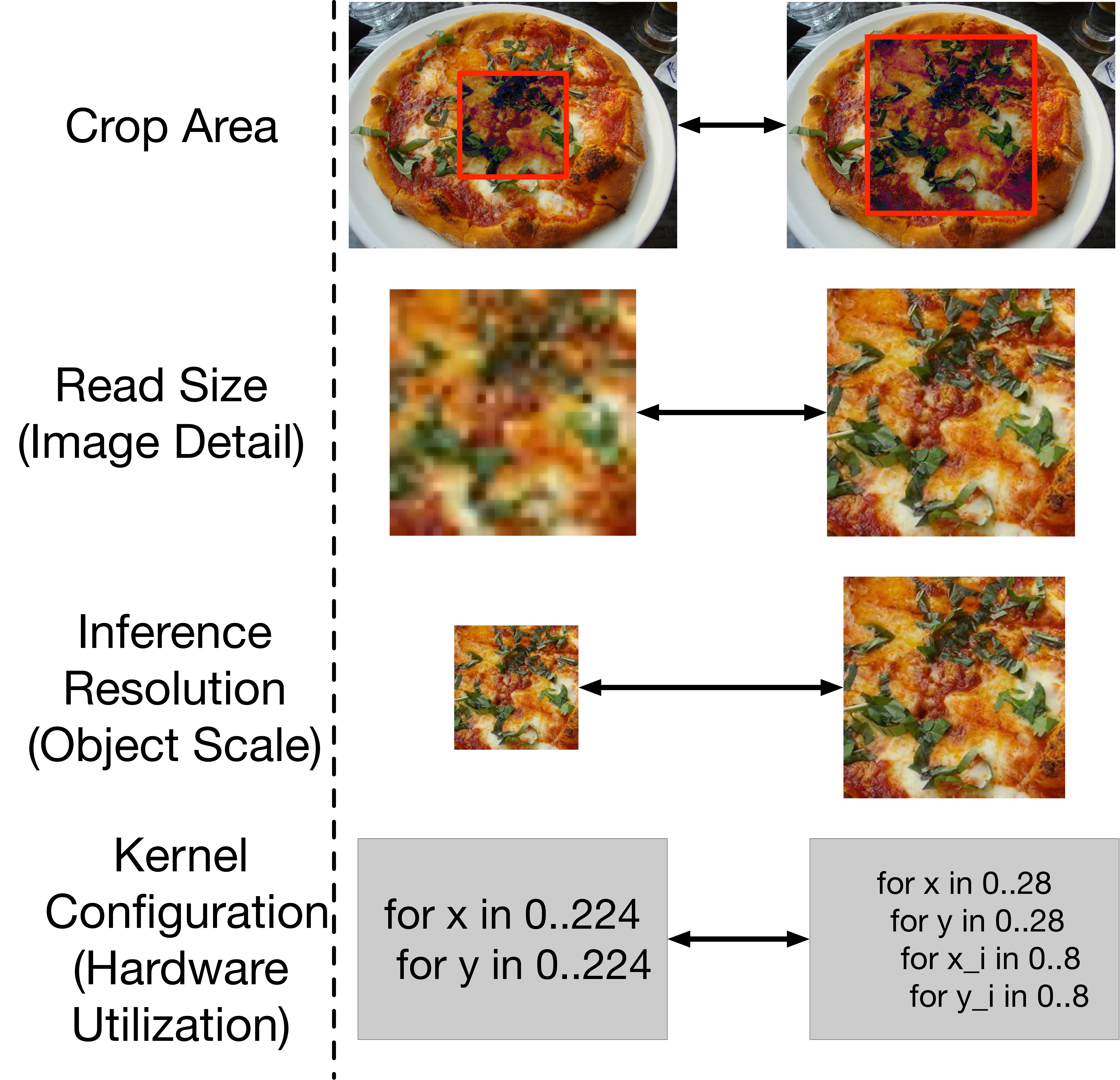}
    \caption{The choice of inference resolution in neural networks introduces associated tightly coupled choices and dependencies, controlling properties such as the apparent size of objects (crop area), image detail (read size), and inference latency (compute kernel configuration). Each choice potentially impacts one or more of model accuracy, inference time, and data storage bandwidth. (Image taken from  ImageNet~\cite{russakovsky2015imagenet}).}
    \label{fig:tradeoffs}
\end{figure}
Modern computer vision models typically run at a fixed image resolution, with this resolution often chosen together with the model architecture to improve model accuracy. In practice, image resolution plays a critical role in the system efficiency of model inference as well. 
The compute throughput requirements scale approximately quadratically with input resolution due to the spatial nature of typical computer vision models and heavy reliance on convolution operations. Storage capacity and bandwidth requirements (bytes moved from storage to compute device per image) also scale quadratically with image resolution, affecting the monetary cost (both storage and network usage are billed) of inference in real-world datacenter or cloud deployments where a separate storage cluster is usually used to store and forward input data through the network~\cite{mudigere2021highperformance}.
As a result, DNN training is frequently dominated by data stall time, which happens both remotely and locally, and can be due to CPU decoding overhead~\cite{mohan2020analyzing}.

While recent work has drawn attention to the importance of proper resolution scaling with respect to algorithmic computational cost (i.e., number of compute operations, or FLOPs) and model accuracy~\cite{tan2019efficientnet}, there lacks a systematic study on the impact of image resolution on accuracy, true wall clock inference latency (FLOPs do not necessarily translate to wall clock time due to different hardware utilization), and storage capacity and bandwidth requirements (images can be read and stored at different resolutions and qualities). 

Thus, we posit that image resolution is an understudied and underexploited hyperparameter in neural network inference. We quantify through a series of measurements and experiments the impact of the choice of different static resolutions and image quality on system resources. Drawing insights from our characterization, we motivate the use of a \textit{dynamic resolution} approach to achieve pareto-optimality in the model accuracy and inference efficiency landscape. 


However, finding the optimal image resolution during inference is challenging.
First, image resolution is tightly coupled to other related parameters, such as crop area and image quality, and has a direct impact on accuracy.
Second, running inference at an image resolution that the underlying framework and numerical libraries are not optimized for results in severely underutilized compute resources, potentially nullifying the savings brought by the choice of a lower resolution. 
Given these dependencies, a holistic approach must be used to determine the best combination in tandem (\autoref{fig:tradeoffs}).

In this work, we begin by characterizing the tradeoff spaces introduced by resolution and describe mechanisms that enable operating at different points in this space.
Following this characterization, we design and implement a two-stage pipeline that uses a lightweight model to select the best inference resolution for a larger backbone model, with the goal of recovering most of the accuracy of choosing the ``correct'' resolution for inference (\autoref{sec:invariance}).
We address the challenge of picking the most efficient image quality by posing it as a calibration task via an image quality metric.
We address the challenge of suboptimal hardware compute utilization when running inference at arbitrary resolutions using operator \emph{autotuning} or choosing an inference implementation by measuring the performance of different implementations that exist in a predefined search space.  
Our contributions include:
\begin{itemize}
\item We propose and characterize methods for dynamic resolution inference and selectively reading image data.
\item We show that resolution-specific kernels improve performance at all resolutions. Particularly,
 tuned inference at $280\times280$ is $1.2\times$ to $1.7\times$ faster than a hardware-specific library implementation at $224\times224$.
\item For image storage, we show that up to 20-30\% of image data can be ignored (reducing storage bandwidth pressure) without losing model accuracy at high resolution inference.
\item We show that a dynamic resolution approach is a viable alternative to fine-tuning to match training and test object scales when accounting for potentially unknown or unfavorable crop sizes.
\end{itemize}

\section{Related Work}
\paragraph{Improving Model Inference Efficiency}
Improving the the computational efficiency of neural network inference is a rapidly evolving field of research due to the high computational costs associated with modern convolutional architectures.
Frequently, these approaches introduce a quality--computational cost or accuracy--computational cost tradeoff space.
To date, most efforts are focused on approaches that only improve \textit{computation efficiency}, such as quantization~\cite{rastegari2016xnor, zhou2016dorefa, fromm2020riptide}, weight pruning~\cite{ji2018tetris, frankle2018lottery}, input masking~\cite{yang2018energy}, temporal redundancy reduction~\cite{buckler2018eva2}, model cascades~\cite{shen2017fast},  alternative numerical representations~\cite{kim2016dynamic, lee2017energy, kalamkar2019study}, image resolution resizing~\cite{touvron2019fixing}, and  memory hierarchy-aware approximation~\cite{lin2021accelerating}.

Equally important but often ignored, storage capacity and bandwidth are precious resources that are closely tied to inference efficiency, as the era of cloud computing has popularized pricing models that meter storage and network usage.
These constraints highlight the importance of providing additional opportunities of improving storage efficiency by minimizing the number of bytes read (or transferred across the network) at inference time, through tuning of the input resolution and quality to computer vision models for accurate inference.

\paragraph{Dynamic Input Resolution}
Image resolution does not need to be a static hyperparameter for model inference because not all classification tasks or categories require the same level of image detail for accurate inference.
Furthermore, image resolution in neural networks is closely tied to the perceived \emph{scale}, or sizes of objects in images.
Recent work~\cite{touvron2019fixing} has pointed out that the choice of resolution implicitly biases the model towards a specific distribution of object scales (the apparent size of objects) based on data augmentation choices at training time.
While a proposed fix~\cite{touvron2019fixing} is to fine-tune the model for the expected distribution of object scales at test-time, this solution relies on the assumption that the test distribution is known and fixed, which is often not the case in real-world scenarios.

On the other hand, the issue of scale dependence is an area of active research with model architecture~\cite{DBLP:journals/corr/KanazawaSJ14}, fine-tuning~\cite{touvron2019fixing}, data augmentation~\cite{hoffer2019mix}, and equivariance-based approaches~\cite{sosnovik2019scaleequivariant} being proposed.
More recently, RS-Nets~\cite{wang2020resolution} have further refined the accuracy vs. scale curve for convolutional neural networks.
However, even with improved robustness or decreased scale dependence, the ideal resolution (and hence the ideal storage fidelity and model operator configurations) for each image may be different.
Unless model accuracy no longer \emph{changes} with scale or resolution, the choice of resolution and the downstream dependencies remain relevant.
Even in the ideal case of fully scale equivariant models, image resolution remains a tunable hyperparameter that dictates the amount of information or fine-grained detail present in images in addition to the size of feature maps.
From this perspective, the mapping of image data to different resolutions and tuning resolution-specific kernels is orthogonal to the motivation behind multi-resolution support.

An important requirement of efficient dynamic resolution support is the availability of high performance kernel implementations for each combination of resolution, model, and hardware. As we later show, running inference at an image resolution that the underlying framework or library is not optimized for causes severe hardware underutilization. But implementing optimal operators for each combination of resolution, model and hardware is impractical. Thus, we rely on work in automatic deep learning kernel optimizations~\cite{ragan2013halide, chen2018tvm, zheng2020ansor, cowan2020automatic} to generate these kernels with minimal programmer effort.

Specializing storage with domain-specific knowledge for increased capacity and performance is also an active area of research~\cite{sampson2014approximate, jevdjic2017approximate, mazumdar2019vignette}.
Prior work has touched on cases where the relative importance of image data (e.g., critical format bits vs. noisy coefficient values) can be matched to storage at different levels of reliability~\cite{guo2016high, jevdjic2017approximate}.

Finally, our proposed dynamic two-model pipeline draws inspiration from Mixture-of-Experts (MoE) approaches in machine learning~\cite{jacobs1991adaptive, shazeer2017outrageously, yang2019soft, lepikhin2020gshard}, where model architectures use a weighted and/or sparse combination of ``experts'' to increase model capacity.
Here, our two-model pipeline can be considered as a modified MoE that uses weight-sharing, with the different experts being the inference resolutions that the backbone model can use.
While prior work has enforced sparse-weighting~\cite{shazeer2017outrageously} or soft-conditioning~\cite{yang2019soft} to efficiently implement the MoE, we use explicit control flow and train the scale and backbone models on separate objectives.

\section{Background}
Efficient support for resolution as a hyperparameter spans storage, algorithm, and computational efficiency considerations. Specifically: 

\begin{itemize}
\item From a storage perspective, we aim to minimize the number of bytes that need to be read and transferred for inference.
\item From an algorithmic perspective, we aim to minimize the number of compute operations (FLOPs).
\item From a computational efficiency perspective, we aim to maximize the utilization of the underlying hardware across difference inference resolutions.
\end{itemize}

In this section, we show how various metrics and factors come into play in the treatment of resolution as a hyperparameter.

\paragraph{Image Quality Metric}
Storage-wise, we focus on the goal of reducing the amount of image data that is read or stored for neural network inference.
One straightforward way to achieve this reduction is to resize large images to lower resolutions, dropping the unnecessary fine details in the images.
However, this approach requires a method to calibrate or map image quality (as a proxy for neural network accuracy) to bytes read from storage. To that end,
image quality metrics such as Peak Signal-to-Noise Ratio (PSNR) and Structural Similarity (SSIM)~\cite{wang2004image} provide relatively fast estimates of image quality given a source or reference image.
One constraint of using image quality metric-guided tuning is that the overhead of computing the image quality metric should be much lower than that of the downstream computer vision model in order to be efficient.
This means that while attractive, expensive image quality metrics such as those that rely on features from neural networks~\cite{zhang2018unreasonable} are too expensive to be practical at this stage in the pipeline.  


\paragraph{Progressive Image Encoding}
The previous discussion made the assumption that reading fewer bytes of image data gracefully degrades image quality, but this assumption is not always true and depends on the image encoding.
To practically use image quality metrics for thresholding to determine the amount of image data to read, we require an image encoding that progressively improves image quality with the amount of data read.
One popular encoding that satisfies this requirement is progressive JPEG, which is a frequency domain-aware data layout that provides a progressive image encoding.\footnote{We use progressive JPEG due to its high degree of compatibility and the ease of transcoding from baseline JPEG; the general approach is orthogonal to the choice of progressive image encoding.}
\begin{figure}
    \centering
    \includegraphics[width=0.45\textwidth]{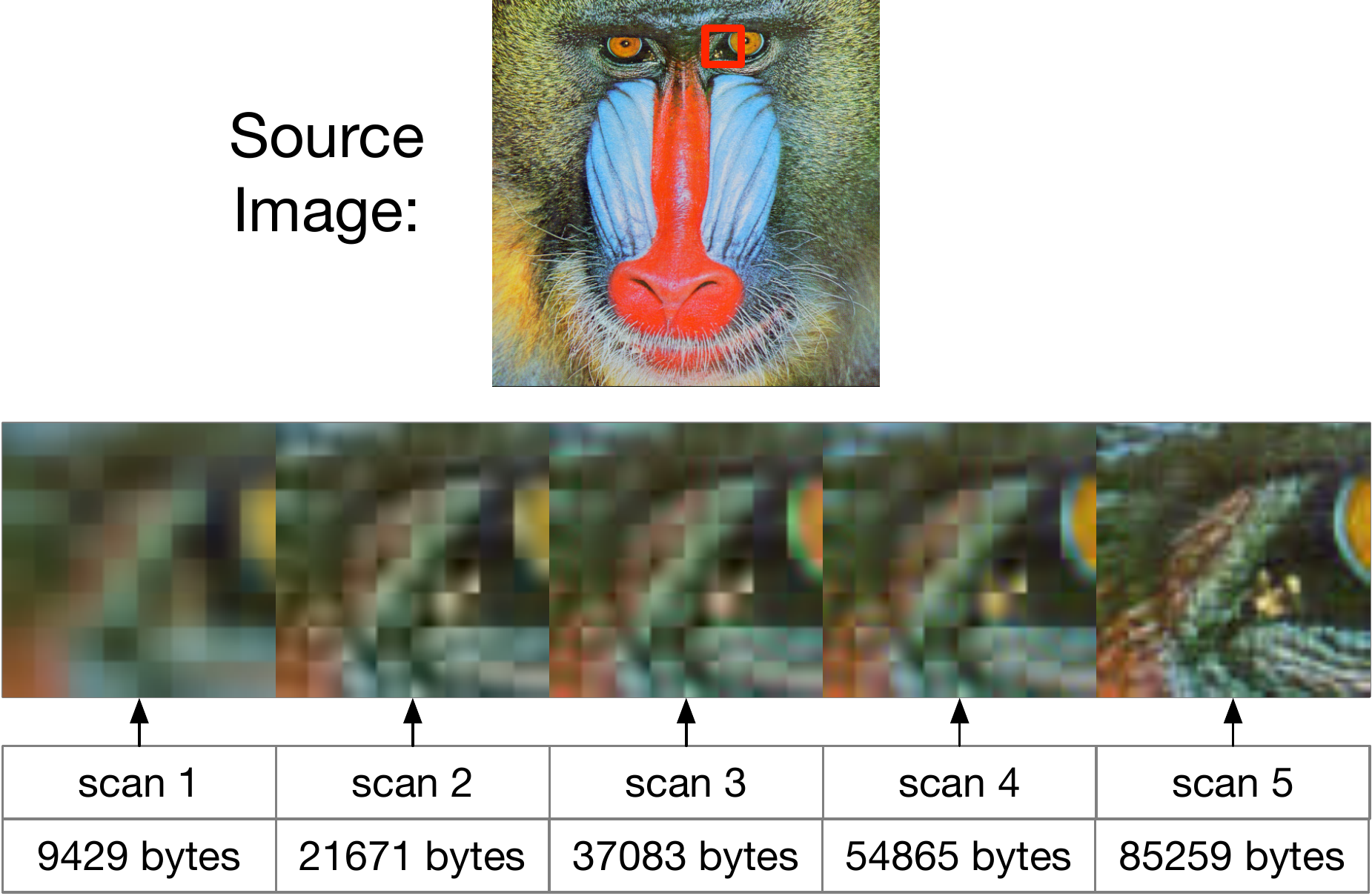}
    \caption{Example of a progressively encoded JPEG image. Each image scan refines previous image data by including higher frequency coefficients. Cumulative bytes read shown below each scan.}
    \label{fig:progressivejpeg}
\end{figure}
Using progressive JPEG, the coarse details of an image are first transferred and rendered, and a lossy preview of the image can be generated before all the image data has been received. \autoref{fig:progressivejpeg} shows an example of how image detail increases as more scans (frequency coefficient groupings in JPEG) of a progressive JPEG image are rendered.
We can use image quality metrics to guide how many scans are needed to reach a target quality for generating lower resolution versions or previews~\cite{yan2017customizing}. 



\paragraph{Crop Sizes, Resolution, and Scale}
\begin{figure}
    \centering
    \includegraphics[width=0.48\textwidth]{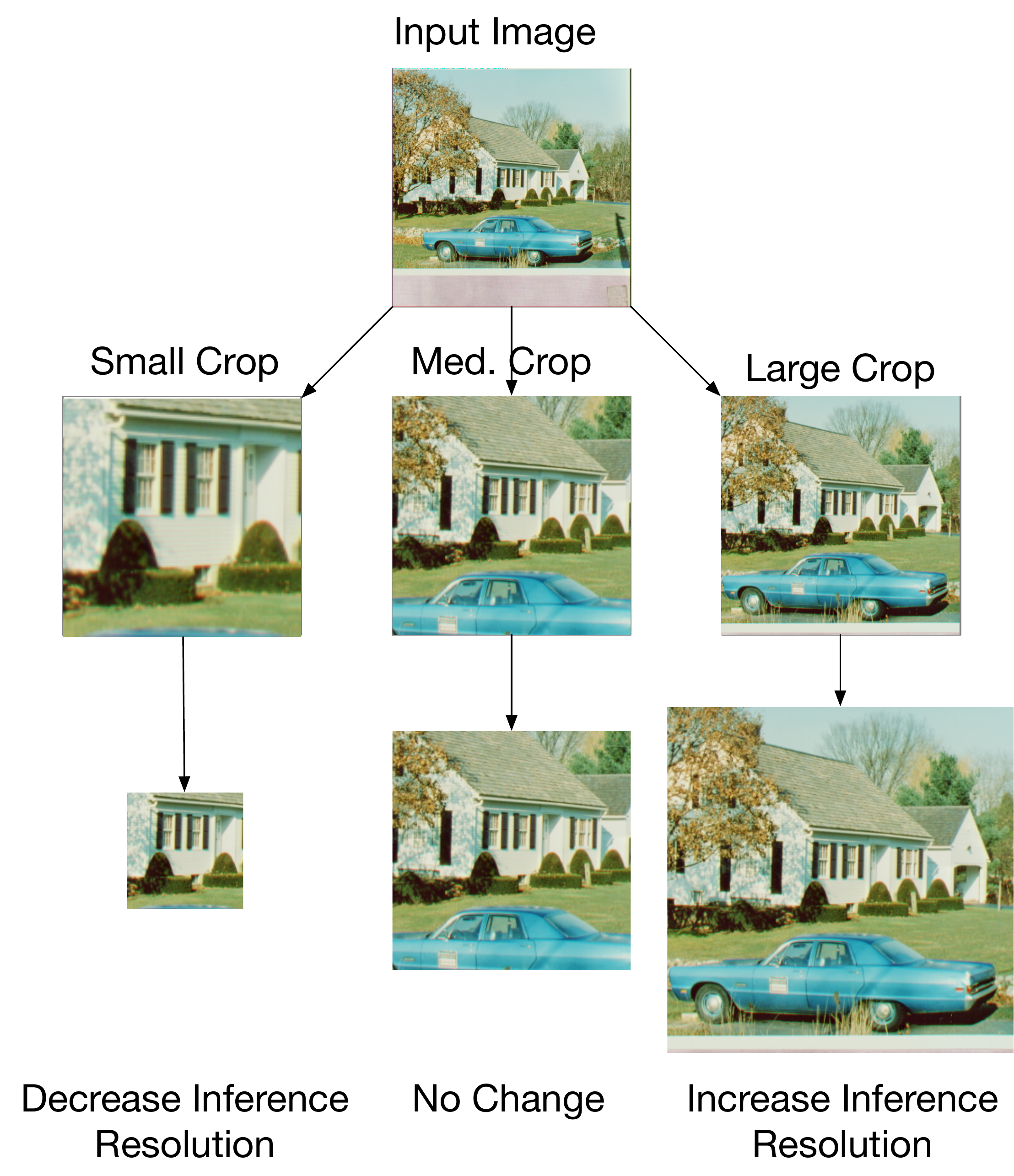}
\caption{Neural networks are sensitive to the apparent scale of objects. We show three different crops of the same image and the corresponding resolution change required to match the object scales across the different crops.}
    \label{fig:cropexample}
\end{figure}
Efficient dynamic resolution support in neural network inference can also be motivated by the lack of scale invariance (or more formally, equivariance)~\cite{sosnovik2019scaleequivariant} in current computer vision models. This issue stems from the fact that while convolution operators are translation equivariant, they are not scale equivariant~\cite{touvron2019fixing}. \autoref{tab:discrepancy} shows an example of resolution/accuracy scaling to illustrate this effect.

\autoref{fig:cropexample} shows an example of how different crop sizes can present objects at different scales to neural networks, and the corresponding change to inference resolution required to compensate for scale differences.
The lack of scale equivariance results in models being sensitive to the distribution of object scales, and even with the typical remedy of data augmentation (e.g., in the form of random cropping), model accuracy can be further improved by fine-tuning on a known scale distribution~\cite{touvron2019fixing}.
We characterize the impact of the issue of popular neural network architectures' \emph{lack} of scale invariance by evaluating model accuracy at several crop sizes; we will see that the favored resolution for model inference heavily depends on the image crop size due to this phenomenon.

\paragraph{Specialized Operator Implementations}
Computationally, reducing image resolution directly reduces the number of floating-point operations (FLOPs) required during inference. But fewer FLOPs does not necessarily ensure lower wall clock inference latency, as potential time savings materialize only when similar compute utilization is achieved when downsampling images.
The compute utilization of deep learning operators is highly dependent on input shapes (e.g., resolution), and we show in~\autoref{sec:eval} that existing libraries do not offer optimized performance for all resolutions, and thus cannot take full advantage of a dynamic resolution scheme. To bridge this gap, we leverage prior work on automatic tensor program optimization~\cite{chen2018learning, ragan2013halide} to generate resolution-specialized operator implementations while minimizing programmer effort.

\section{Choosing Image Resolution via Object Scale}
\label{sec:invariance}
\begin{figure}
    \centering
    \includegraphics[width=0.5\textwidth]{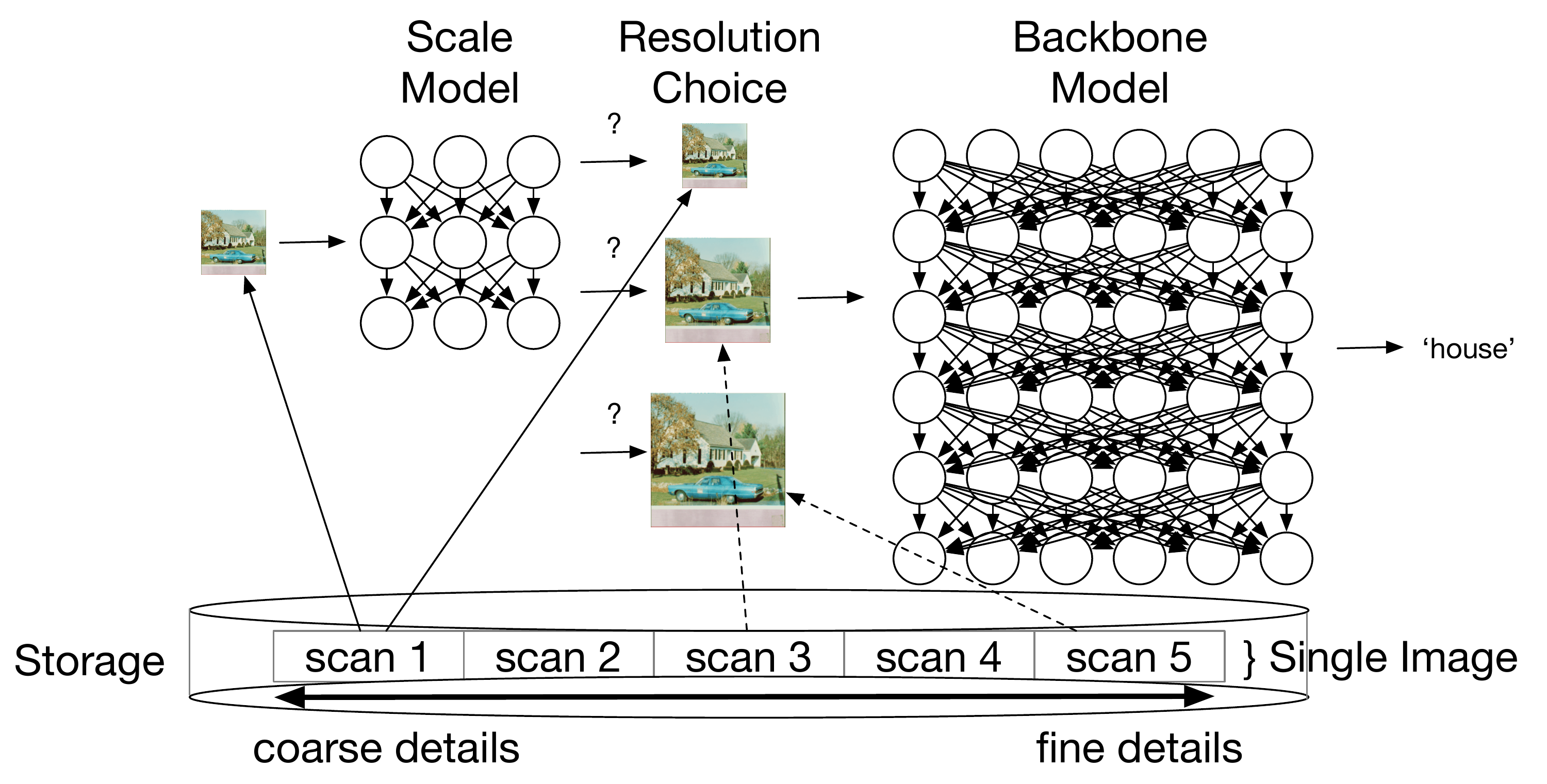}
    \caption{Example of a dynamic resolution system: images are stored with a progressive encoding that arranges each image as a sequence of scans. Low resolution images (in this case $112\times112$) are first sent to a small scale model that predicts the best resolution for inference. If necessary, additional image data is read to produce the appropriate resolution version for inference. }
    \label{fig:pipeline}
\end{figure}

We now describe a simple yet effective way to leverage a dynamic resolution-enabled inference pipeline with object scale-awareness by using two neural network models in series (\autoref{fig:pipeline}) such that the optimal resolution is chosen automatically for each image. We refer to the first model as the \emph{scale} model, as it roughly attempts to predict the appropriate resolution to normalize scale for neural network inference, and the second model as the \emph{backbone} model, as it performs the specified computer vision task at the chosen resolution.
While this two-model pipeline introduces an additional control flow decision (which resolution to choose), this decision is at a coarse granularity (an entire image inference).

\paragraph{Scale Model}
The scale model is trained with a multilabel classification objective (specifically, we use binary cross-entropy loss): it aims to predict whether a trained backbone model will be correct at a given resolution for an input image.
At inference time, we select the resolution chosen by the scale model with the highest predicted likelihood of yielding a correct prediction (from the backbone model).
As higher resolutions only improve accuracy until the scale of objects becomes too small~\cite{touvron2019fixing}, the scale model does not require any modification to the multilabel objective to avoid always choosing the highest resolution.
Since determining object scale does not require fine image details, the scale model can be lower in resolution (e.g., $112\times112$). To further reduce the computational cost of the scale model, we can use a model with an efficient architecture (e.g., MobileNet).

\paragraph{Backbone Model}
The backbone model for each dataset split is a standard classification model without additional modification.
Note that we do not train separate backbones for each resolution and instead rely on  input shape agnostic architectures such as ResNet to reduce training costs.
Running the model at a different resolution than that it was trained with does not degrade accuracy, provided the object scales are matched.
However, the inference cost in terms of FLOPs increases nearly quadratically with the resolution of the backbone model, as most of the cost comes from convolutions.
True wall clock scaling is usually better, as higher compute complexity tends to also increase the utilization of hardware execution.

\begin{table}[!t]
    \normalsize
    \caption{Example of compute complexity scaling with input resolution (in billions of floating-point operations). Here, the accuracy values are obtained by performing inference on a model trained at $224\times224$ resolution, indicating the train-test resolution discrepancy~\cite{touvron2019fixing} where higher resolutions do not always improve accuracy due to a scale mismatch. }
    \centering
    \begin{tabular}{c|c|c|c}
       Model  & Resolution & GFLOPs & Accuracy  \\
    \hline
    ResNet-18 & $112\times112$ & 0.5 & 47.8\\ 
    ResNet-18 & $168\times168$ & 1.1 & 62.7\\ 
    ResNet-18 & $224\times224$ & 1.8 & 69.5\\ 
    ResNet-18 & $280\times280$ & 2.9 & \textbf{70.7}\\ 
    ResNet-18 & $336\times336$ & 4.2 & 70.1\\ 
    ResNet-18 & $392\times392$ & 5.8 & 69.4\\ 
    ResNet-18 & $448\times448$ & 7.3 & 68.9\\ 
    \end{tabular}

    \label{tab:discrepancy}
\end{table}

\begin{figure}
    \centering
    \includegraphics[width=0.48\textwidth]{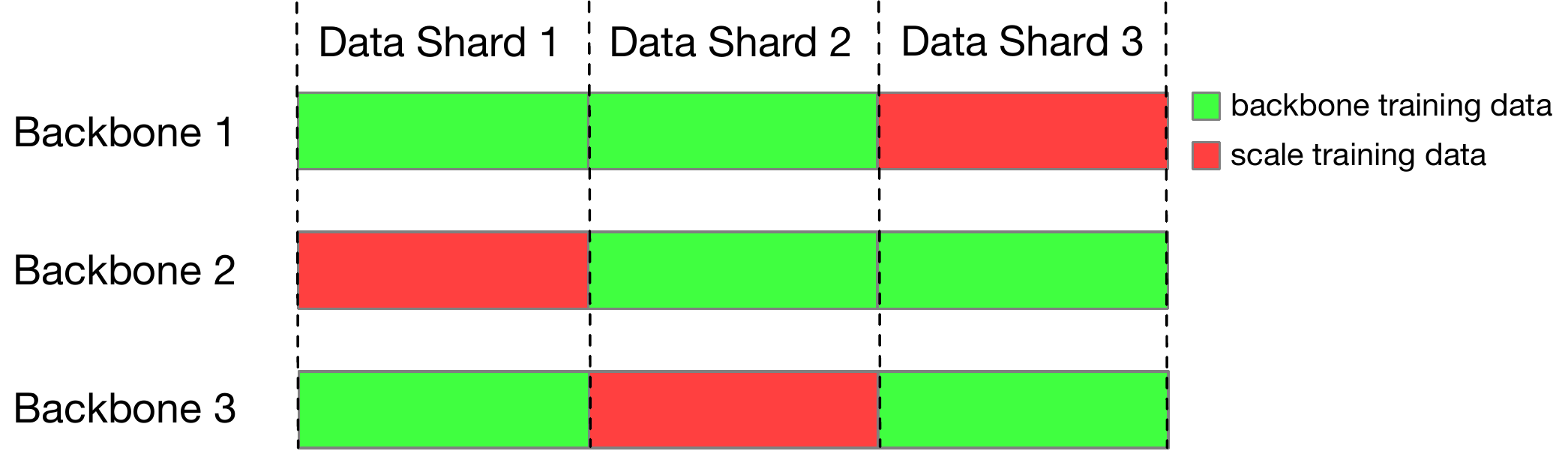}
    \caption{Multiple backbone models are trained on disjoint shards of the training set, and the scale model is trained by alternating backbone models, using the training set shard the current backbone model has not seen.}
    \label{fig:crossval}
\end{figure}

\paragraph{Training}
The multilabel classification objective introduces the problem of another data split, as training the scale model requires an already trained backbone model.
To most efficiently leverage all available data when training the scale model, we train it using a cross-validation style approach (\autoref{fig:crossval}).
Several backbone models are trained on disjoint shards of the training set, and the scale model is trained by alternating between backbone models, using the corresponding training set shard that the current backbone model has not seen (\autoref{fig:crossval}).
For our evaluation, we train four different backbone models on 3/4ths of the ImageNet~\cite{russakovsky2015imagenet} and Cars~\cite{krause20133d} datasets, and train the scale model using the corresponding 1/4th held-out slice for each backbone.\footnote{While we used this setup for evaluation purposes, we have found that the scale model requires considerably less data and fewer training epochs to converge, and can be trained on a single dataset split without much (if any) loss in accuracy to save training time.}
When measuring end-to-end accuracy, we use a backbone trained on the full training set.


\section{Choosing How Much Data to Read}
\begin{figure*}[!t]
    \centering
    \begin{tabular}{@{}c@{}}
    \includegraphics[width=0.497\textwidth]{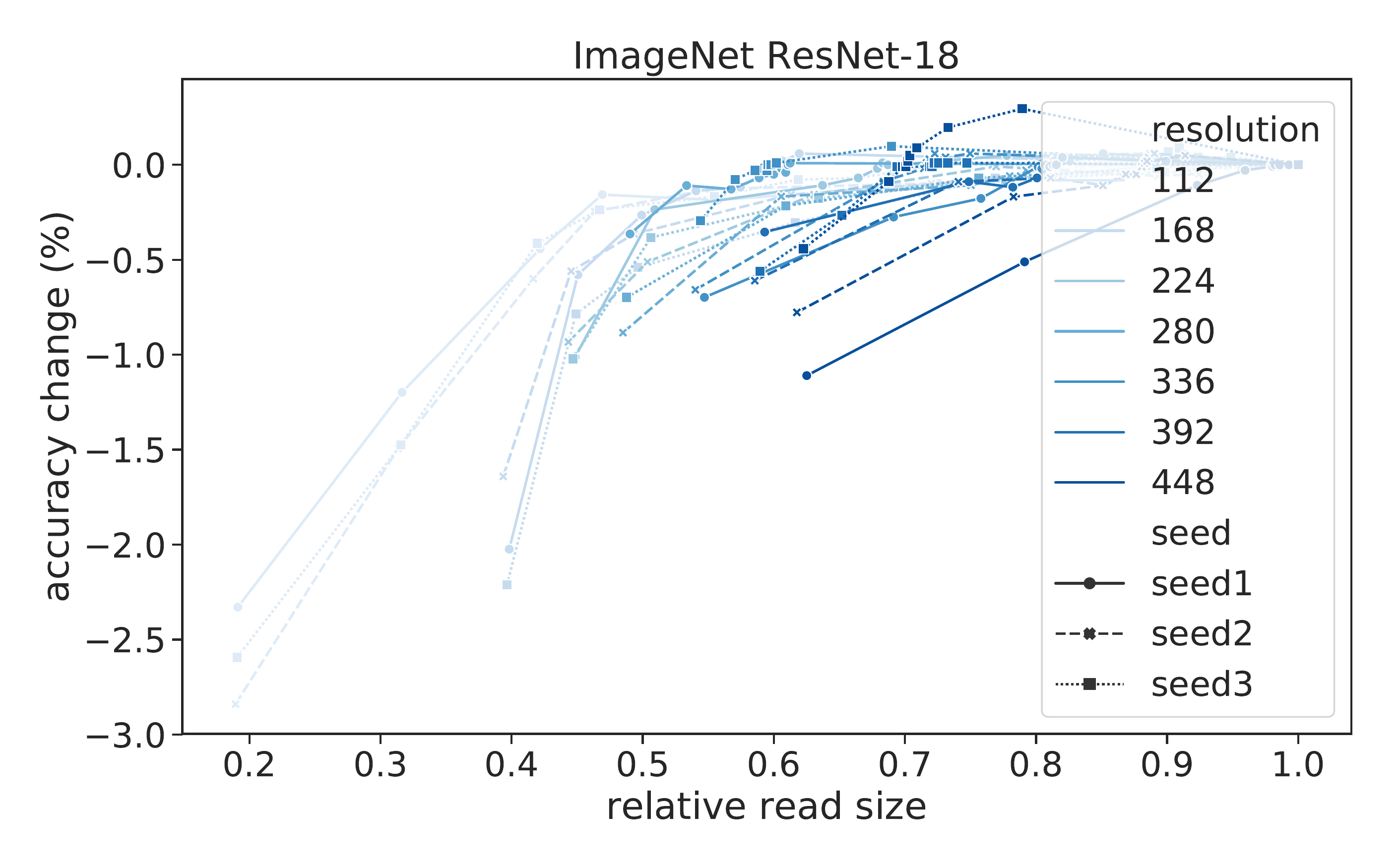}\\
     (a)
    \end{tabular}
    \begin{tabular}{@{}c@{}}
    \includegraphics[width=0.497\textwidth]{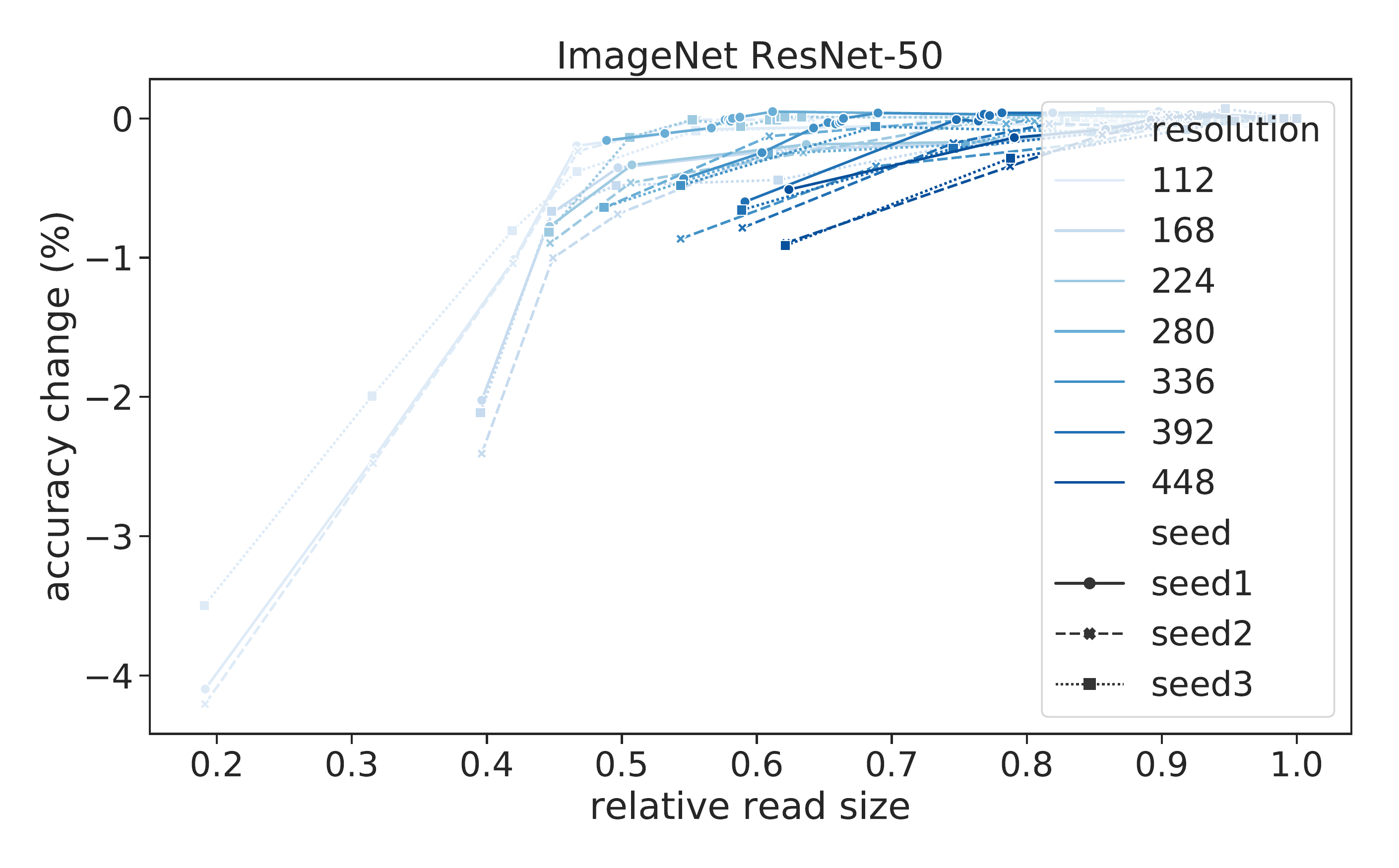}\\
     (b)
    \end{tabular}
    
    \centering
    \begin{tabular}{@{}c@{}}
    \includegraphics[width=0.497\textwidth]{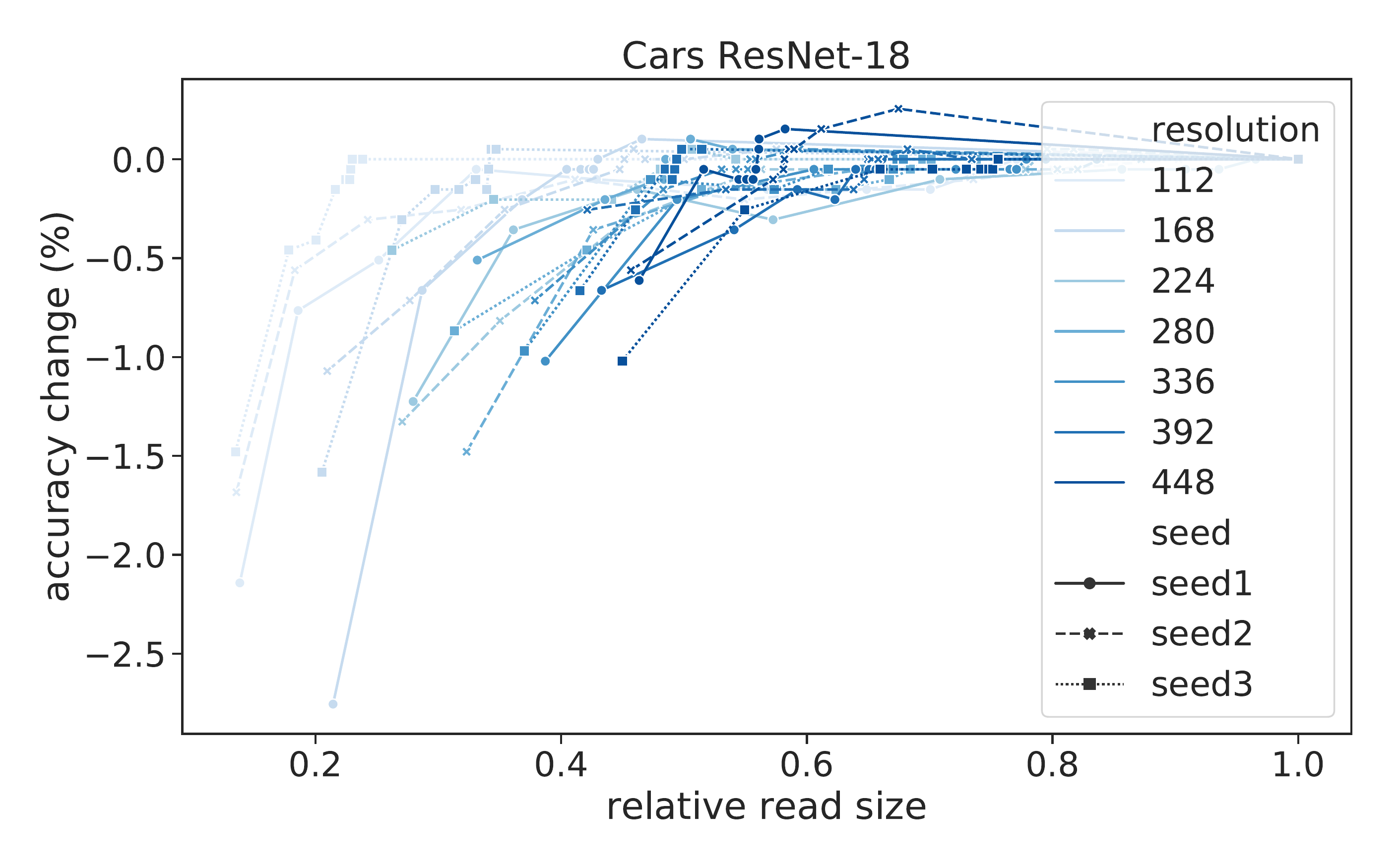}\\
    (c)
    \end{tabular}
    \begin{tabular}{@{}c@{}}
    \includegraphics[width=0.497\textwidth]{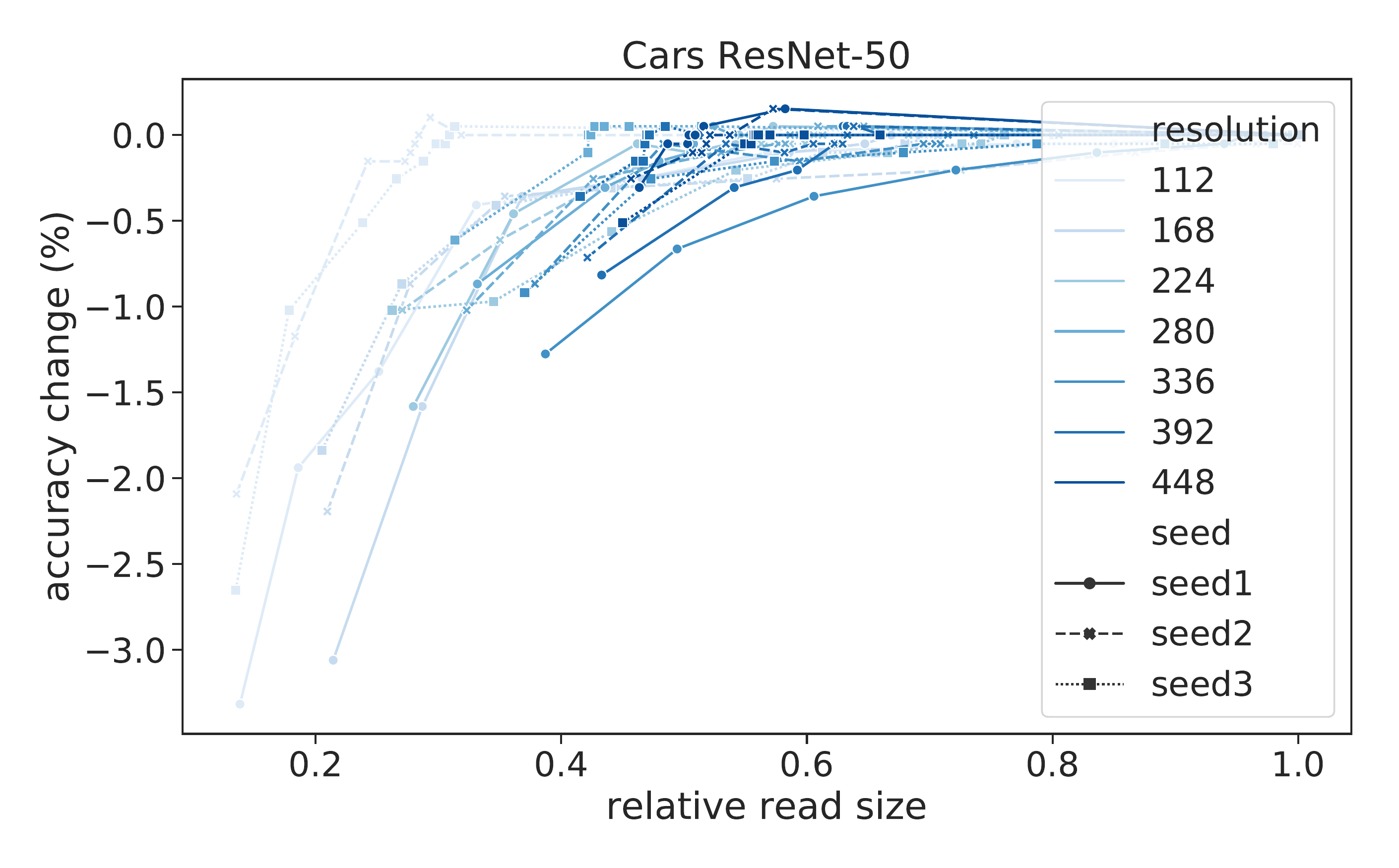}\\
    (d)
    \end{tabular}
    \caption{Storage calibration: relative (to reading the entirety of image data) top-1 accuracy change of ResNet-18 and ResNet-50 on ImageNet (a, b) and Cars (c,d) at different resolutions with varying amounts of image data read. The three seeds indicate different models trained and evaluated with different train/test splits. 
    The amount of image data read (1.0 indicates the entire file is read) is determined by sweeping a range of SSIM values and progressive JPEG scans. Lower resolutions require less image data for the same SSIM value, but accuracy degrades more rapidly with respect to the amount of image data read.}
    \label{fig:calibration_cars_ssim}
\end{figure*}

Besides image resolution, image quality is another important parameter to tune to reduce the quantity of image data read while maintaining accuracy. Naturally, the amount of data required for accurate inference is data-dependent and lacks a one-size-fits-all value. Thus, we pose the problem of finding the optimal image quality as a storage \emph{calibration} task using an image quality metric. In this calibration task, we use a small amount of training data to tune quality thresholds to determine the sufficient amount of image data to read to maintain accuracy.
Concretely, given a quality metric, such as SSIM, we calibrate thresholds from model accuracy vs. quality levels (\autoref{fig:calibration_cars_ssim}) for each resolution.
These curves can form the basis for a storage policy that chooses the amount of data to read for a given resolution requested by a computer vision model.

We explain in detail how this calibration works using the same datasets (ImageNet and Stanford Cars) as examples, as they differ in resolution distribution and type of classification and illustrate the generality of the approach.
While the Cars dataset contains fewer images, some are of considerably higher resolution than the ImageNet dataset: the average dimensions of training images in the Cars dataset are $699\times482$ pixels while the average dimensions of images in the ImageNet dataset are $472\times405$ pixels.

To search for the minimal quality (SSIM threshold) that satisfies an accuracy target (e.g., within 0.05\% accuracy loss), we run binary search over the SSIM interval [0.94, 1.0], and terminate the search after the step size falls below 0.0001, with the constraint that no more than 0.05\% accuracy is lost for each of the resolutions.
We use three different train/validation splits of Cars and ImageNet (shown as the different seeds in \autoref{fig:calibration_cars_ssim}), and use just 10,000 images for calibration per split due to time constraints. 

\autoref{fig:calibration_cars_ssim} shows accuracy vs. the relative amount of data read (normalized to 1.0, corresponding to reading all image data), averaged over a collection of images in the training set of ImageNet and Cars respectively.
Overall, lower resolutions require less image data for the same SSIM value, but accuracy degrades more rapidly with respect to the amount of image data saved compared to higher resolutions.
The curves of accuracy vs. image read size appear to be shifted left for Cars vs. ImageNet: accuracy is better preserved even when images are loaded at low fidelity for Cars.
This difference supports our claim that dataset differences require separate image quality threshold tuning given an accuracy target, because image features have different levels of importance in different datasets (e.g., abstract shapes may be more important in Cars vs. fine-grained textures in ImageNet).

On the other hand, one trend common to both datasets is that higher image resolutions often required lower image quality (when compared to the ground truth resized image) to maintain model accuracy.
This trend is perhaps surprising, because intuitively, one might expect a benefit of higher input resolution to be the inclusion of details that are lost in lower resolution images,  requiring \emph{higher} quality. In fact, this trend is pronounced enough that maintaining accuracy at higher inference resolutions may require \textit{less} image data to be read than for inference at lower resolutions: for Cars, minimal accuracy losses were observed at higher resolutions even when just over half the image data was read.

%

\section{Maximizing Hardware Utilization}
Common computationally intensive operators such as convolution typically require highly specialized implementations on hardware such as GPGPUs or even CPUs with vector instructions to achieve good utilization.
However, most libraries today only provide optimized implementations for the most commonly used shapes/resolutions, and thus the benefits of storage and algorithmic savings may be overshadowed by inefficient operator implementations. 
We now discuss how to find the optimal implementation for every resolution. 

To provide specialized implementations of operators for different image resolutions, one must consider hardware details such as the organization and size of the compute units (e.g., CUDA cores), memory hierarchy (e.g., shared/scratchpad memory, caches, and global DRAM), bus bandwidth, and input sizes and data layouts.
These dependencies mean that implementations are also highly sensitive to input shapes, which are coupled to the input resolution of the neural network model.
Thus, manual implementation of these specialized operators for each resolution and input shape is highly impractical, especially when considering the wide variety of model architectures and operators. 

Fortunately, we can leverage prior work on automatic compiler optimizations~\cite{chen2018learning} for shape-specific deep learning kernels to generate a specialized implementation for each resolution. 
Automatic operator tuning (autotuning) models the search process as a black box optimization problem to find the best combination of parameters including loop tiling factors, data layouts, and loop orderings among others, driven by direct measurement of  end-to-end performance.
With autotuning, we can characterize and close the throughput gap (\autoref{fig:tuning}) between high and low resolution inference, especially among library implementations that may be overfitted to specific resolutions.
Although we focus our discussion on optimizations for inference using commodity CPUs, the autotuning process has been successfully applied to various hardware device types including GPUs.


\section{Evaluation}
\label{sec:eval}

We now provide a comprehensive evaluation of the effects of resolution on inference accuracy and efficiency.
Specifically, we show (1) how operator autotuning bridges the gap of hardware efficiency when running inference with a dynamic resolution; (2) how we use dynamic resolution to better reduce FLOPs without losing as much inference accuracy when compared to a static approach; (3) how we can use image quality threshold calibration to further reduce storage bandwidth requirements without a large impact on accuracy.


\paragraph{Closing the Throughput Gap for Dynamic Resolution}
\label{sec:tuning}
\begin{figure*}[tb]
    \centering
    \begin{tabular}{@{}c@{}}
    \includegraphics[width=0.245\textwidth]{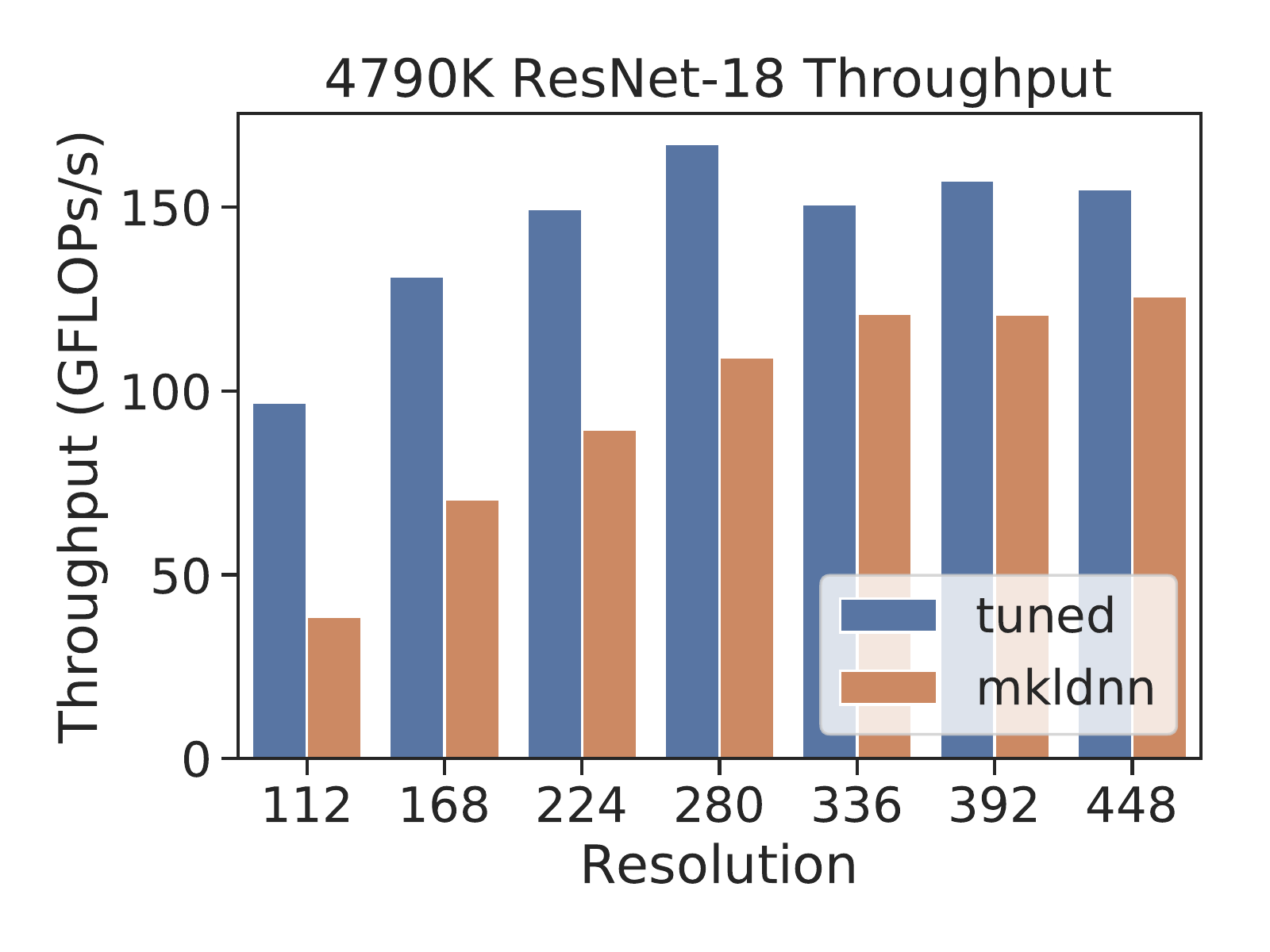}\\
     (a)
    \end{tabular}
    \begin{tabular}{@{}c@{}}
    \includegraphics[width=0.245\textwidth]{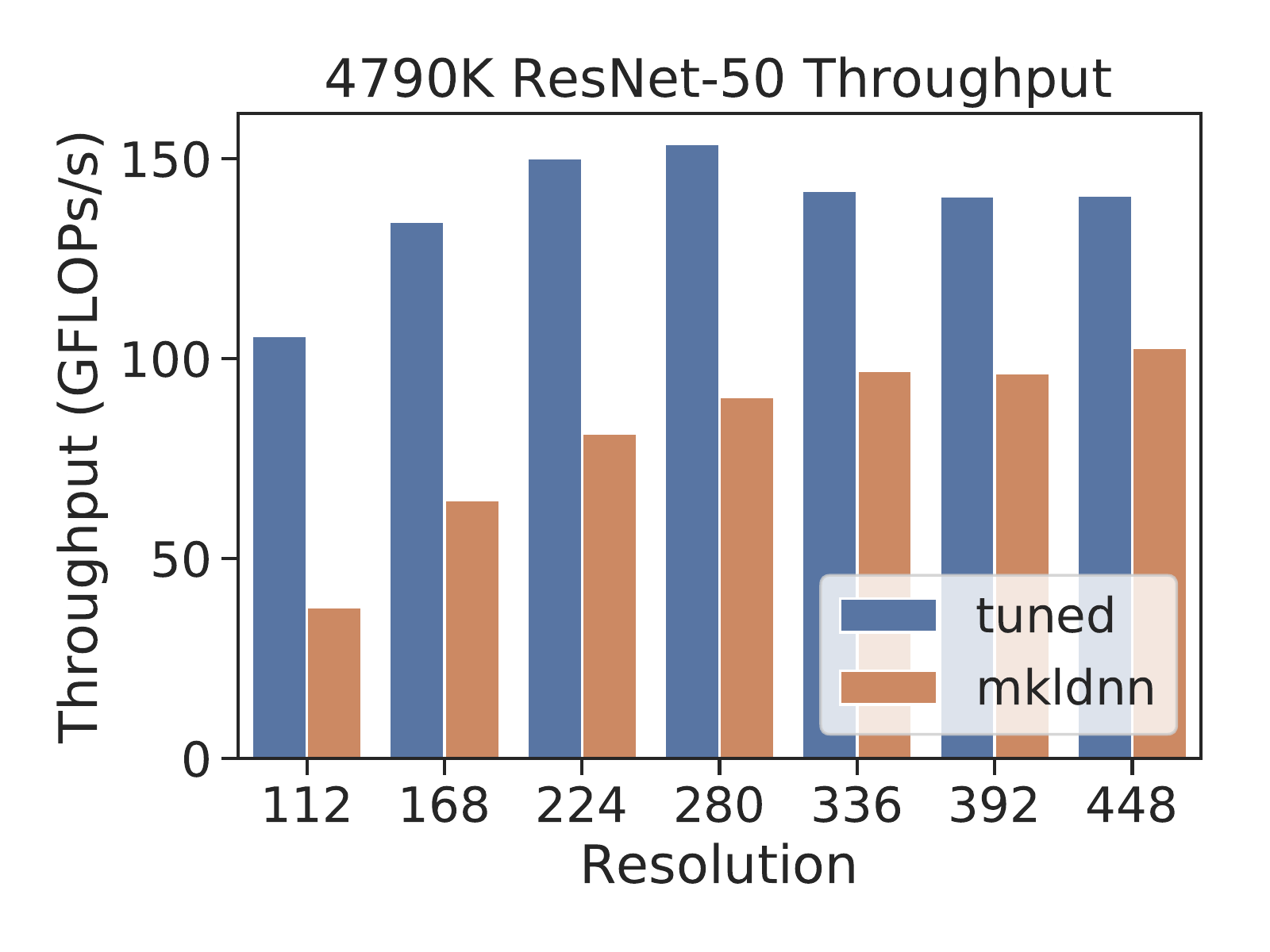}\\
     (b)
    \end{tabular}
    \begin{tabular}{@{}c@{}}
    \includegraphics[width=0.245\textwidth]{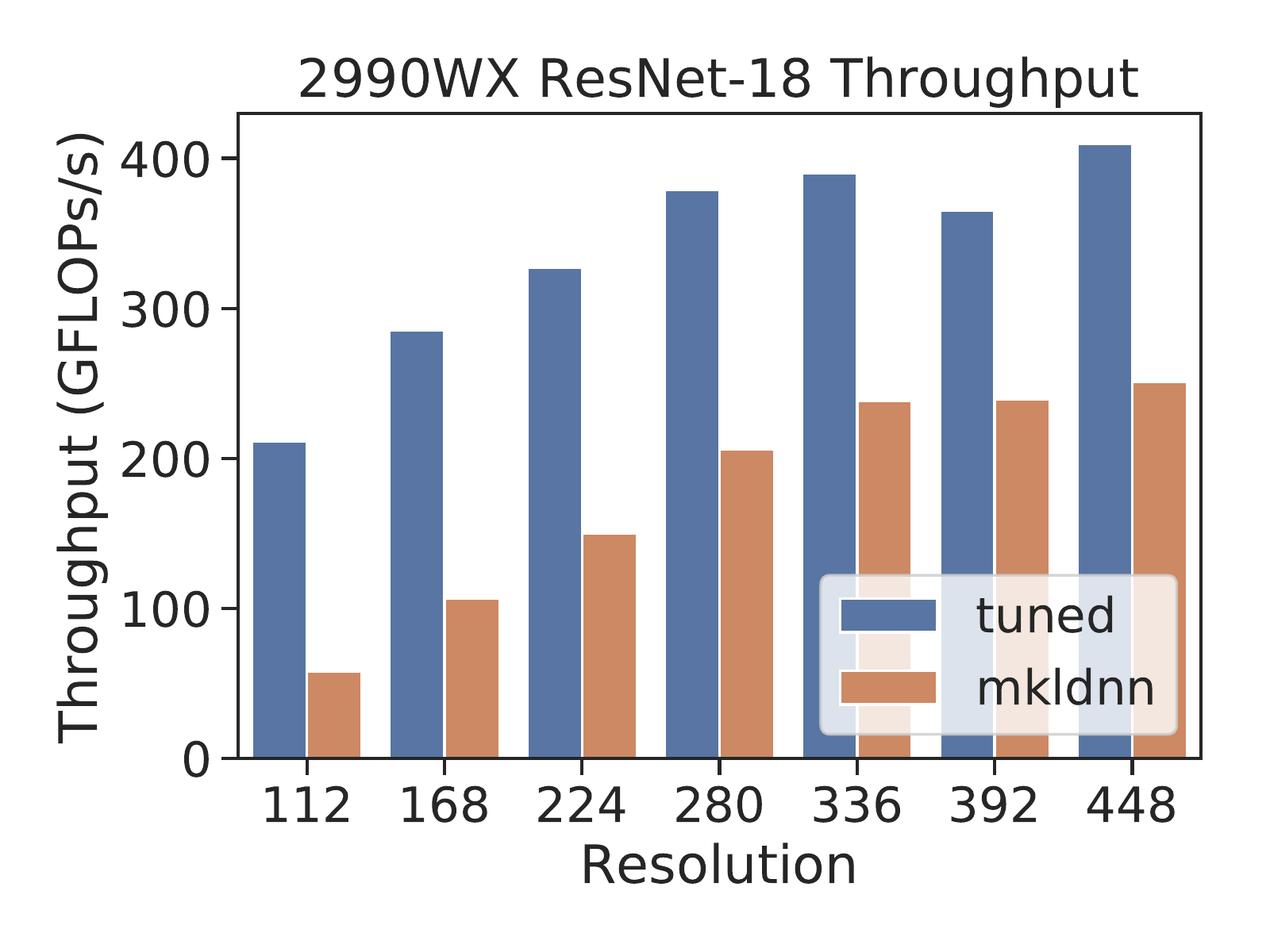}\\
     (c)
    \end{tabular}
    \begin{tabular}{@{}c@{}}
    \includegraphics[width=0.245\textwidth]{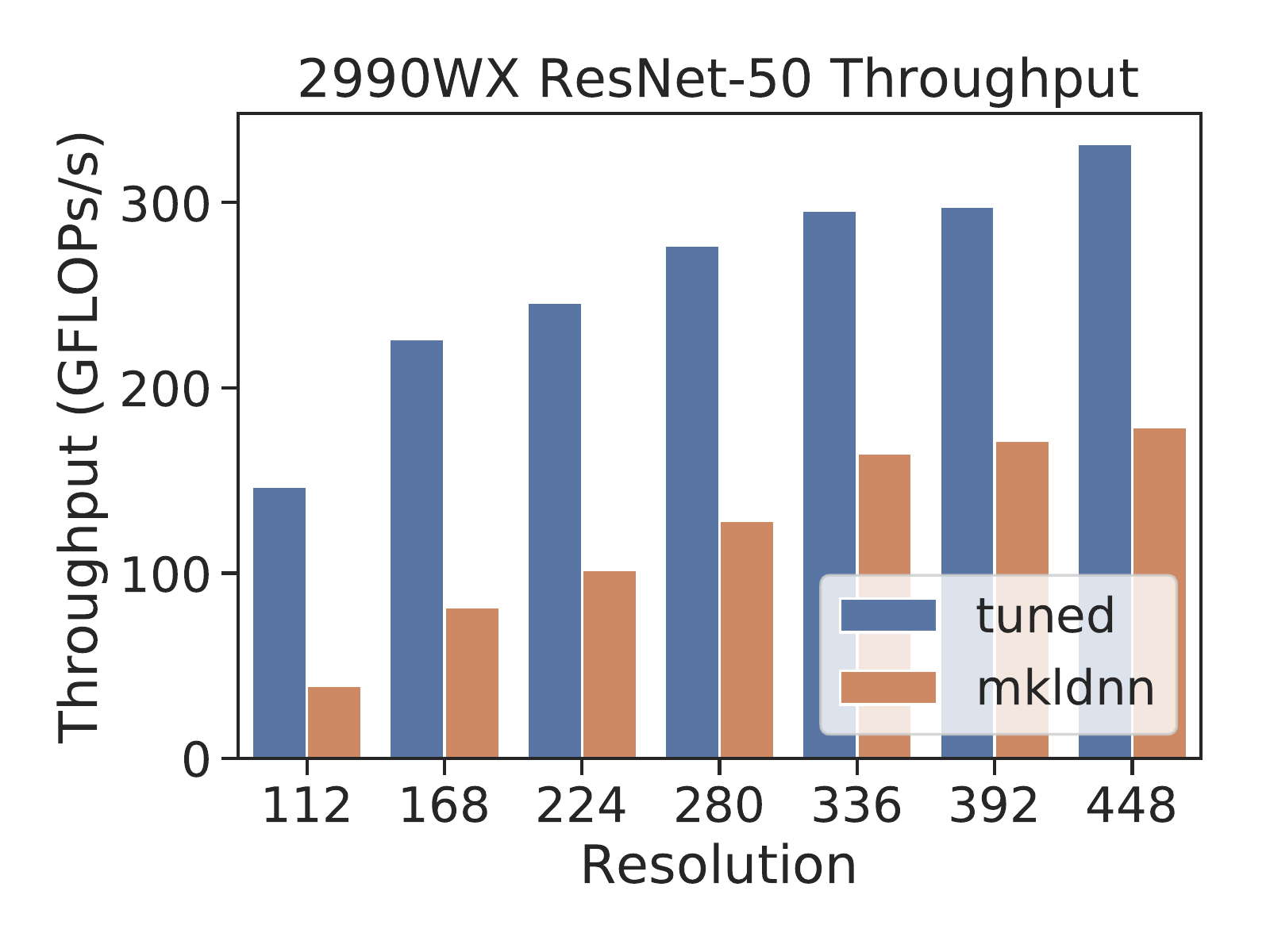}\\
     (d)
    \end{tabular}
    \caption{Throughput of ResNet-18 and ResNet-50 at different resolutions using the Intel MKLDNN Library compared with tuned implementations measured on Intel 4790K and AMD 2990WX. Tuning achieves higher throughput/utilization at lower resolutions.}
    \label{fig:tuning}
\end{figure*}
We run autotuning for convolution layers, the most heavyweight operator type used in ResNets on the Intel 4790K and AMD 2990WX CPUs.
We compare the tuned version with one backed by the  operator implementations provided by Intel MKLDNN via PyTorch, a widely used, state of the art operator library and deep learning framework.
~\autoref{fig:tuning} compares the inference throughput using Intel MKLDNN with the specialized version for each resolution under the typical inference scenario of batch size one, showing that the raw throughput can be doubled through autotuning. \autoref{tbl:wallclock} compares the wallclock time of ResNet-50 using tuned and library implementations, showing that higher resolutions can be be used without slowdown (relative to the library implementation) with tuning.
We use typical thread counts for CPUs with simultaneous multithreading (SMT), where \emph{half} the total number of hardware threads are used.\footnote{We observed in some cases that using fewer than half the number of hardware threads could increase performance! As tweaking the number of hardware threads constitutes tuning in and of itself, we report performance for the typical configuration.}
Beside inference latency improvement, the tuned implementations achieve better throughput (even compared to tuned high resolution kernels) for lower resolutions that have fewer FLOPs on both the 2990WX and 4790K (\autoref{fig:tuning}).

\begin{table}
\normalsize
\centering
\begin{tabular}{c | c c c c}
     & \multicolumn{2}{c}{4790K} & \multicolumn{2}{c}{2990WX} \\
     Res & Tuned & MKLDNN & Tuned  & MKLDNN  \\
     \hline
     112 & 10.3 & 28.8 & 7.4 & 27.6\\
     168 & 18.9 & 39.1 & 11.2 & 31.0\\
     224 & 27.6 & 50.9 & 16.8 & 40.7\\
     280 & 43.4 & 73.7 & 24.1 & 51.8\\
     336 & 66.6 & 97.6 & 32.0 & 57.4\\
     392 & 93.4 & 136.1 & 44.1 & 76.6\\
     448 & 117.5 & 161.1 & 49.9 & 92.5\\
\end{tabular}
\caption{Wallclock latency (ms) of ResNet-50 with tuned and library implementations on 4790K (4 Cores) and 2990WX (32 Cores).
When increasing resolution, tuning enables higher resolution inference without slowdown relative to the library implementation.
When decreasing resolution, tuning better sustains throughput relative to the library implementation.}
\label{tbl:wallclock}
\end{table}


We draw two conclusions from these findings that establish the feasibility of dynamic resolution by improving end to end inference latency through autotuning:

(1) \textit{Tuned operators achieve peak CPU FLOPs/s with a smaller resolution}. This means through tuning, throughput has much less reliance on high resolution images. In fact, we find the ideal operating point to be $168\times168$ due to the lower compute complexity while most of the throughput of higher resolutions is also attainable with tuning. 

(2) \textit{Autotuning can better materialize latency reduction by sustaining high throughput in low resolution inference}. By reducing resolution from $448\times448$ to $112\times112$, with ResNet-18 and ResNet-50, we expect the ideal speedup to be around $15\times$, assuming equal hardware utilization. However, only $ 4.6 \times$ and $ 5.6 \times$ speedup (on the Intel platform, and $ 3.5  \times$ and $ 3.3 \times$ on the AMD platform) is realized by the library implementation. Through autotuning, the achieved speedup rises to $ 9.4  \times$ and $ 11.4  \times$ (on the Intel platform, and $ 7.7 \times$ and $ 6.7 \times$ for the AMD platform) respectively.

\label{sec:accflops}
\begin{figure*}[!ht]
    \centering
    \begin{tabular}{@{}c@{}}
    \includegraphics[width=0.245\textwidth]{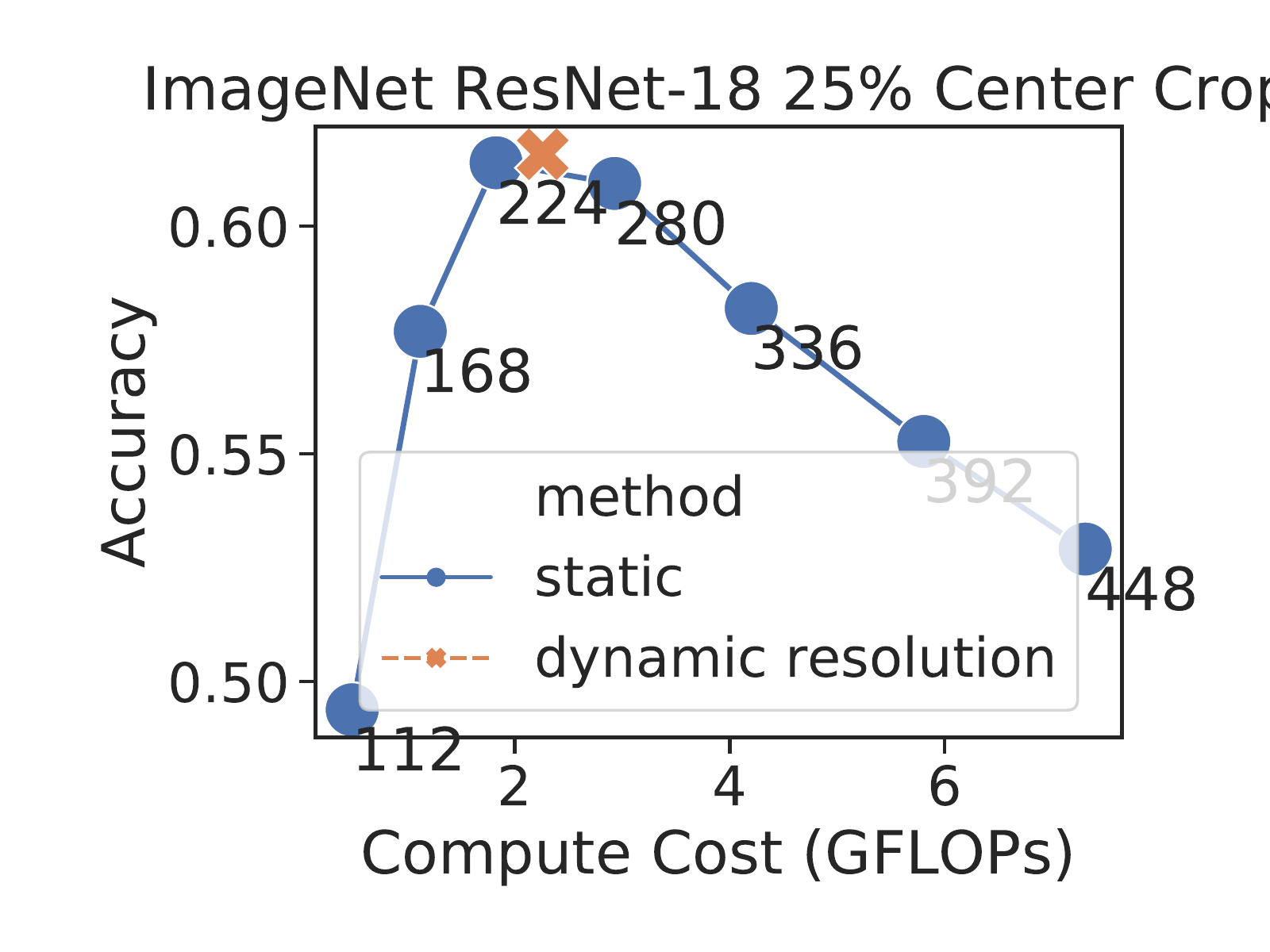} \\
     (a)
    \end{tabular}
    \begin{tabular}{@{}c@{}}
    \includegraphics[width=0.245\textwidth]{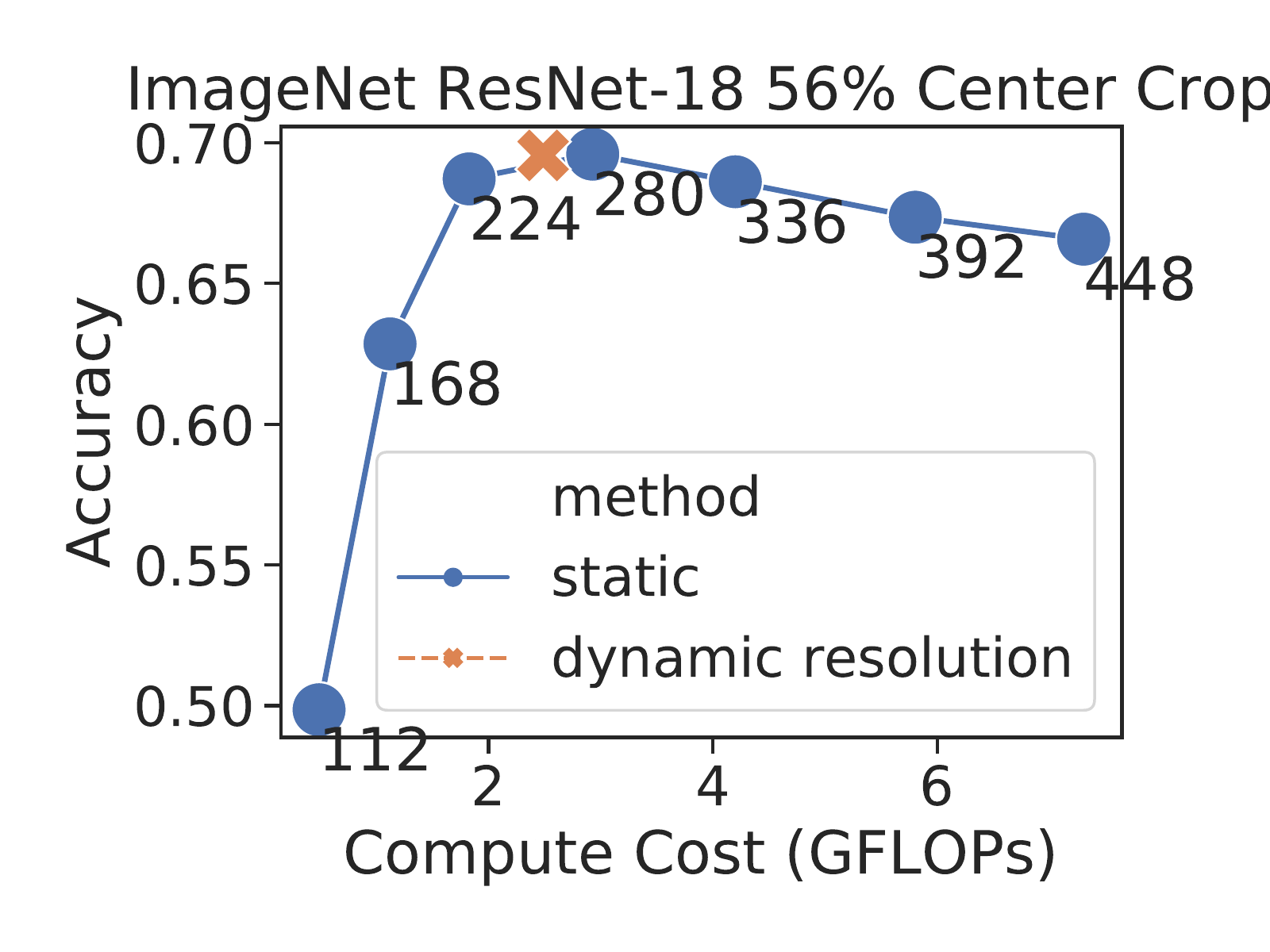} \\
     (b)
    \end{tabular}
    \begin{tabular}{@{}c@{}}
    \includegraphics[width=0.245\textwidth]{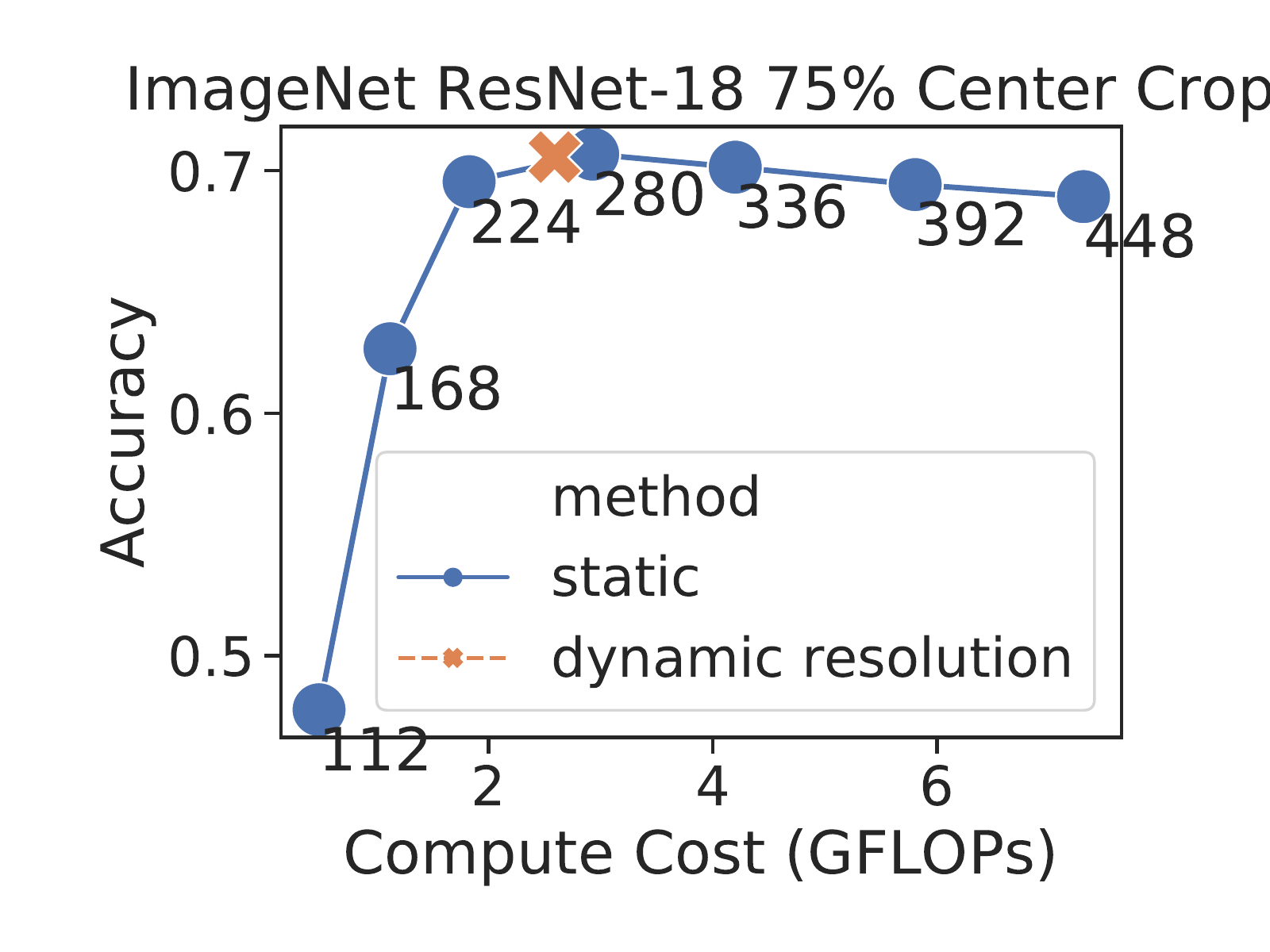} \\
     (c)
    \end{tabular}
    \begin{tabular}{@{}c@{}}
    \includegraphics[width=0.245\textwidth]{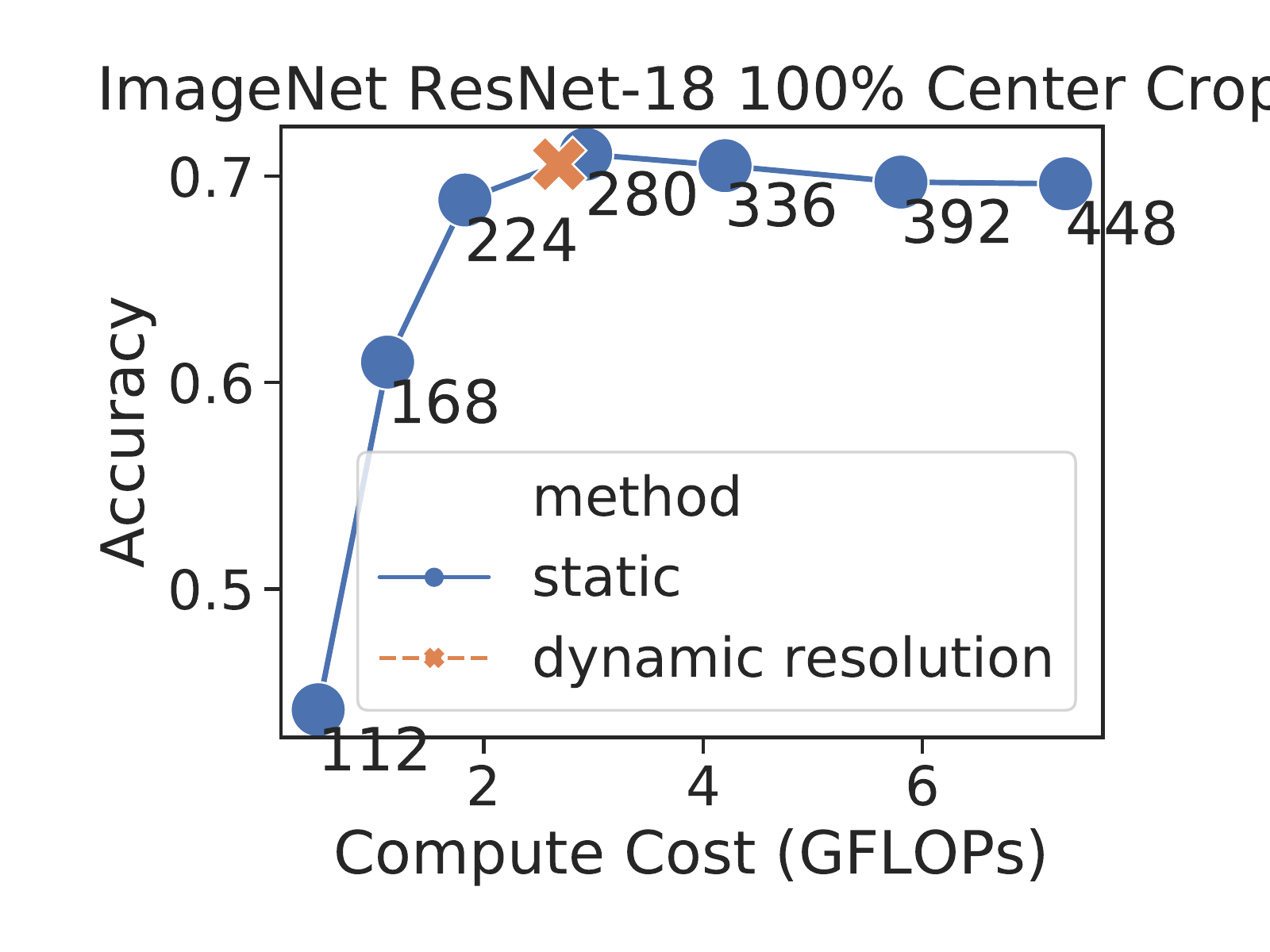} \\
     (d)
    \end{tabular}
    \begin{tabular}{@{}c@{}}
    \includegraphics[width=0.245\textwidth]{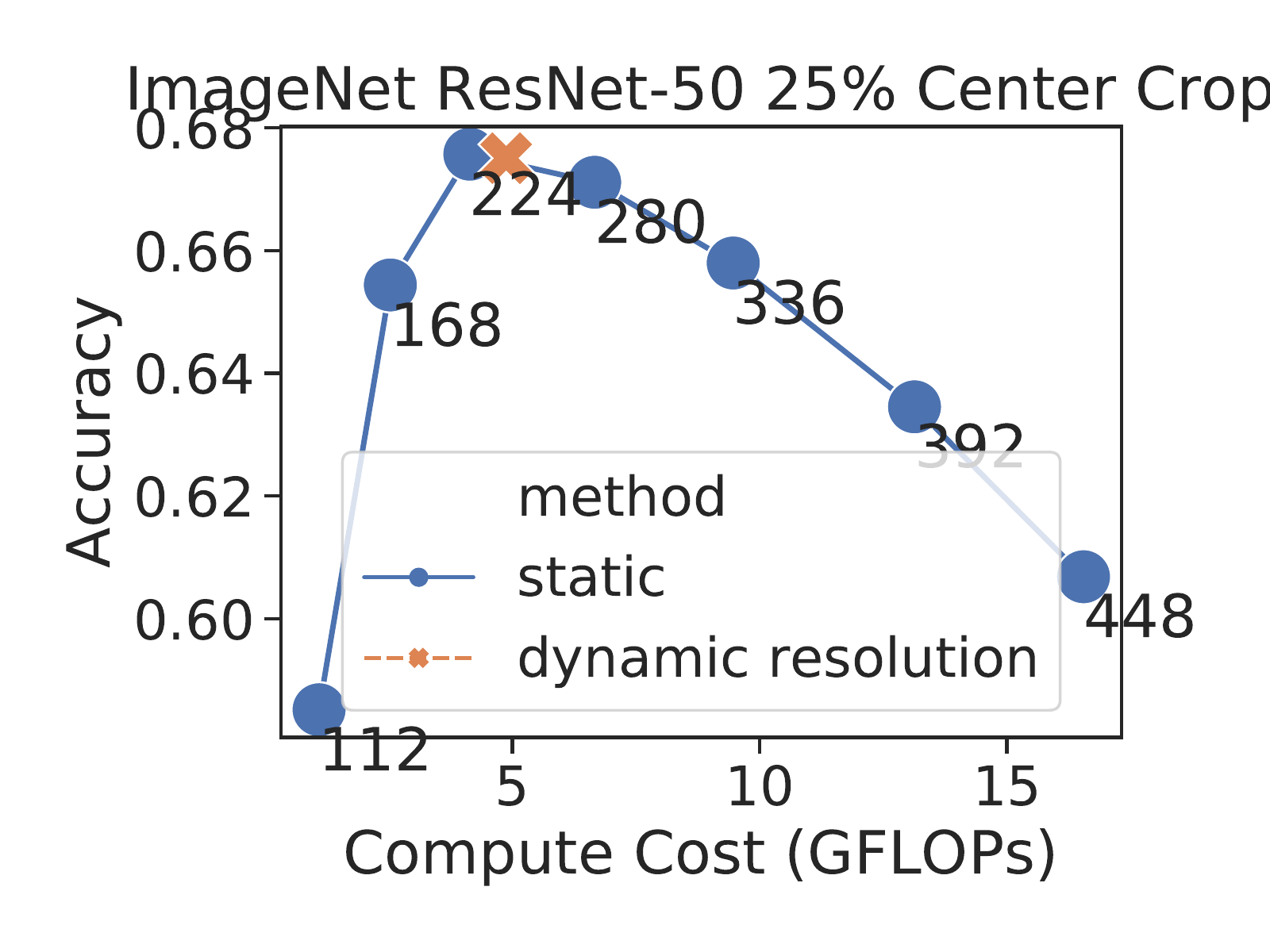} \\
     (e)
    \end{tabular}
    \begin{tabular}{@{}c@{}}
    \includegraphics[width=0.245\textwidth]{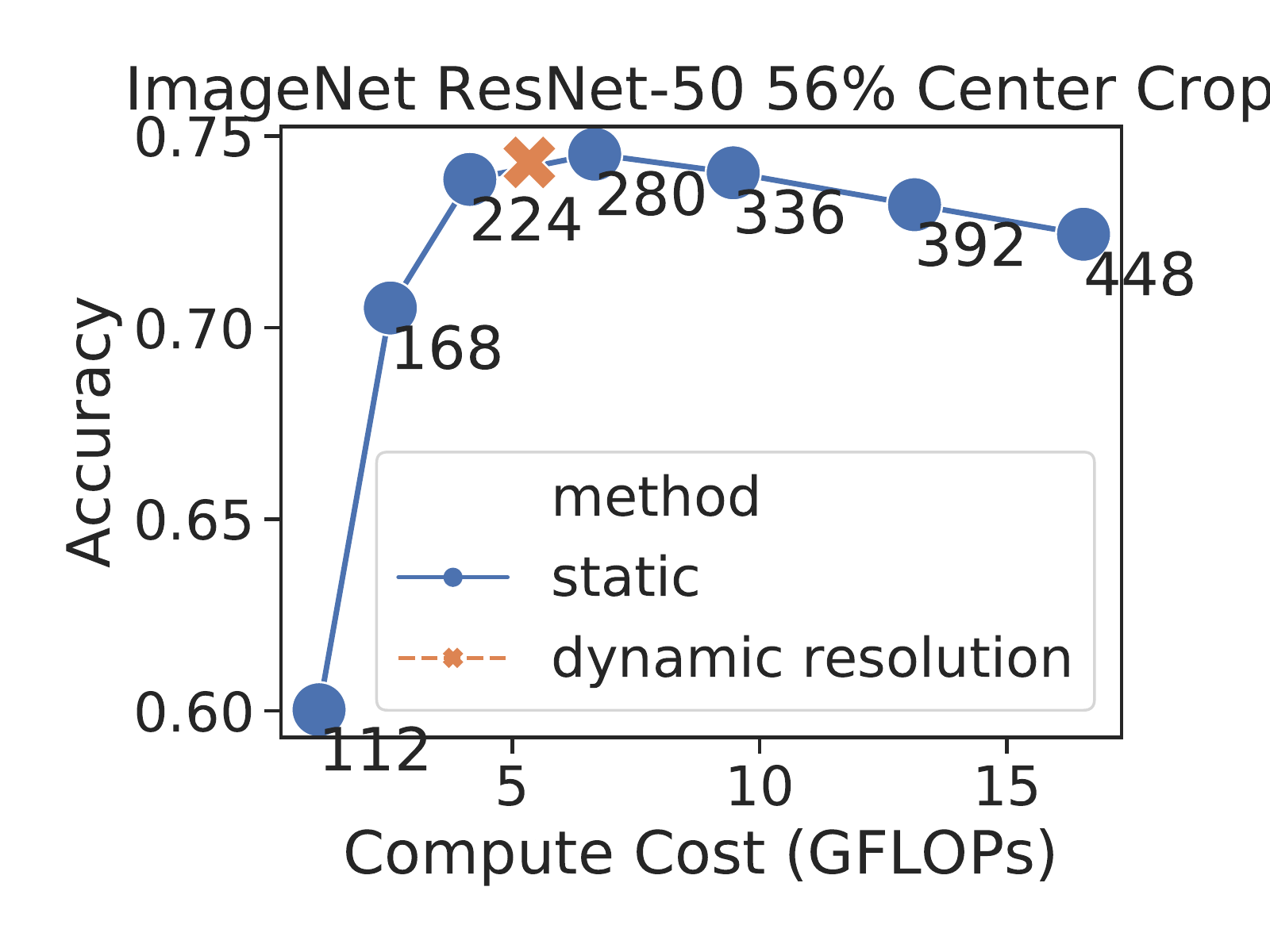} \\
    \small (f)
    \end{tabular}
    \begin{tabular}{@{}c@{}}
    \includegraphics[width=0.245\textwidth]{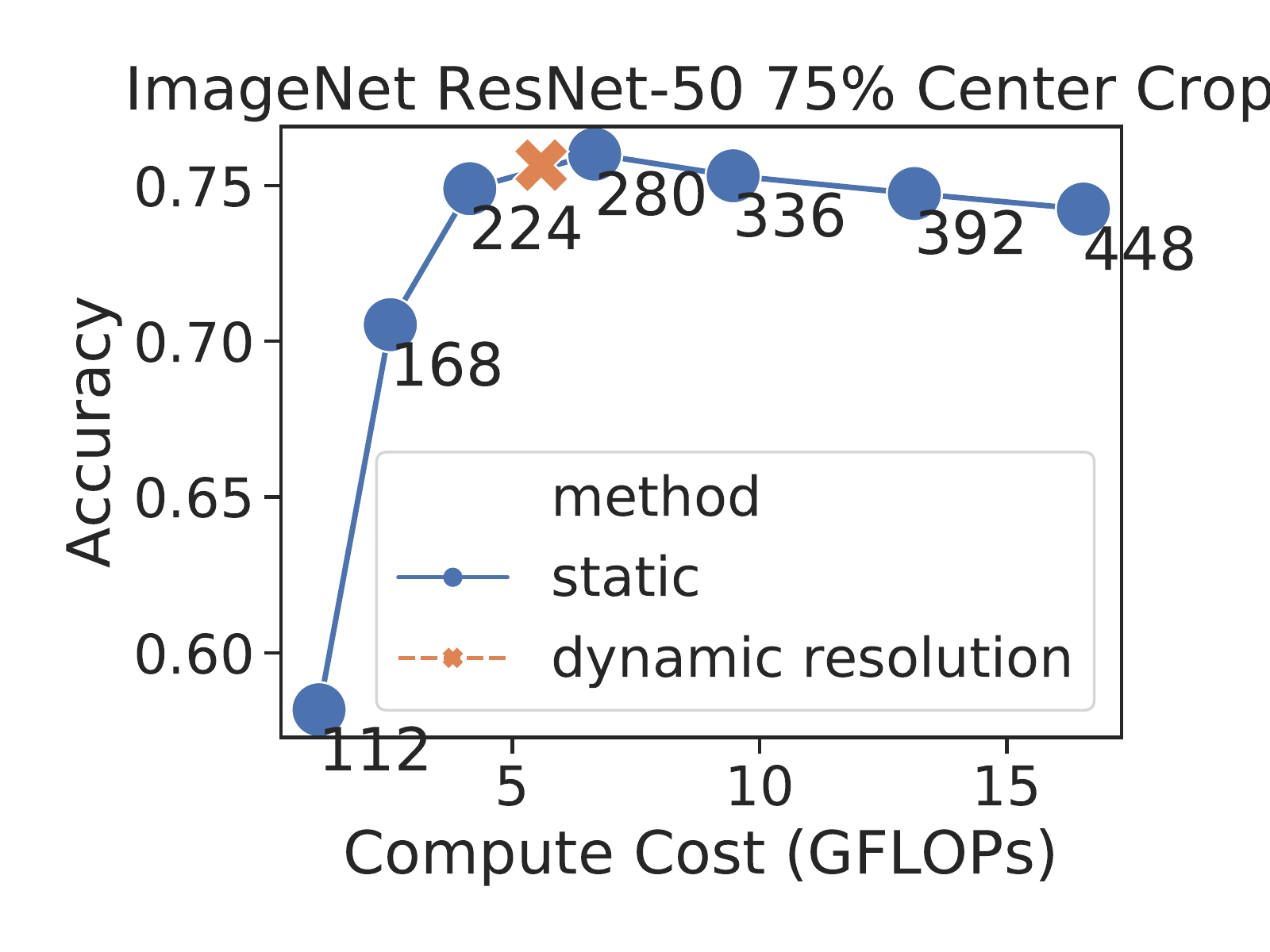} \\
    \small (g)
    \end{tabular}
    \begin{tabular}{@{}c@{}}
    \includegraphics[width=0.245\textwidth]{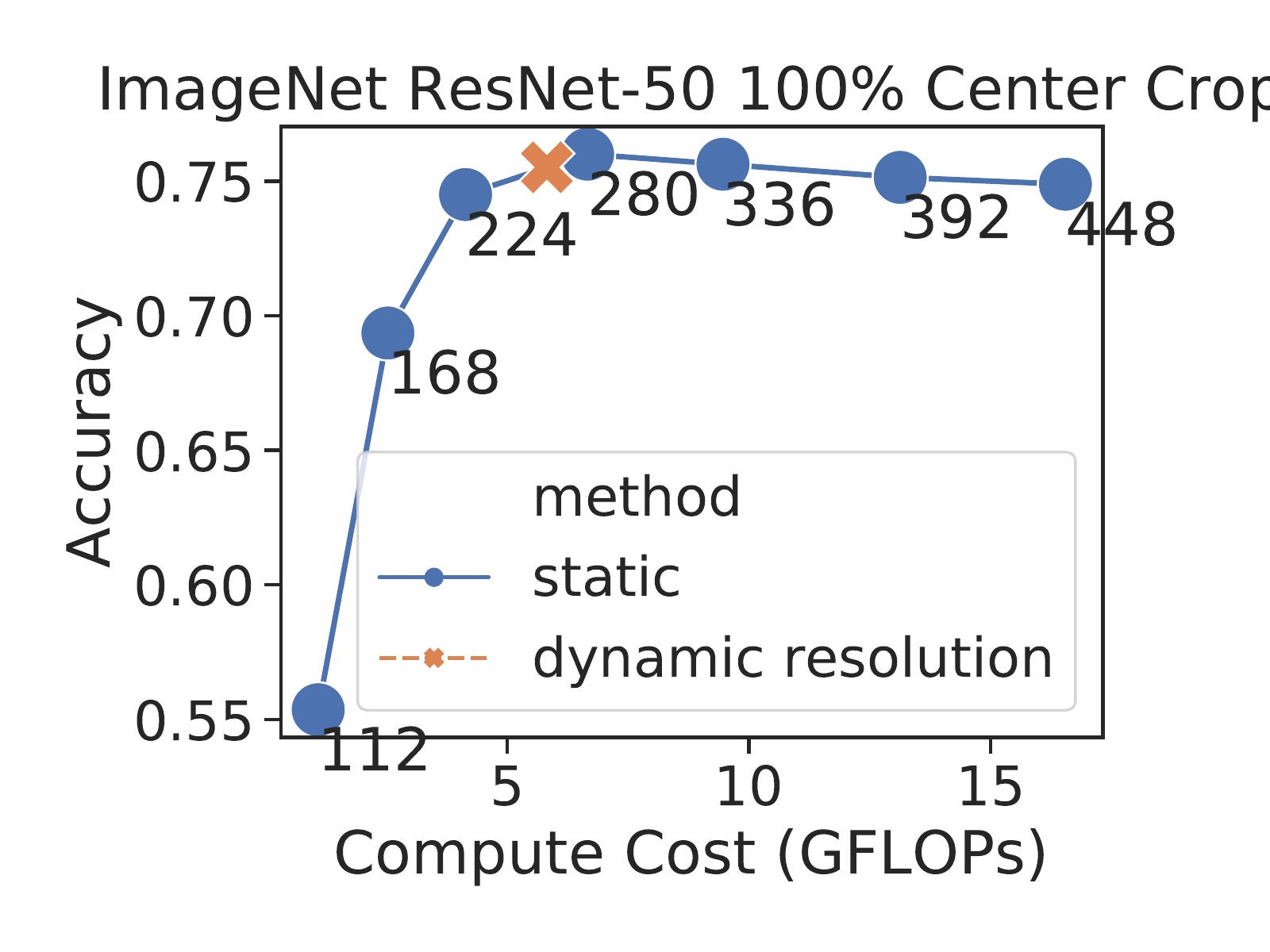} \\
    \small (h)
    \end{tabular}
    \caption{Accuracy vs. average FLOPs with static and dynamic resolution using ResNet-18 (a-d)/50 (e-h) on ImageNet. Crop sizes increase left to right from 25-100\%. Smaller crops favor lower resolutions while larger crops favor higher resolutions due to the models' dependence on object scale. The dynamic resolution approach operates near the apex of each curve, without a predefined, static resolution.}
    \label{fig:accflops_resnet_imagenet}
\end{figure*}
\begin{figure*}[!ht]
    \centering
    \begin{tabular}{@{}c@{}}
    \includegraphics[width=0.245\textwidth]{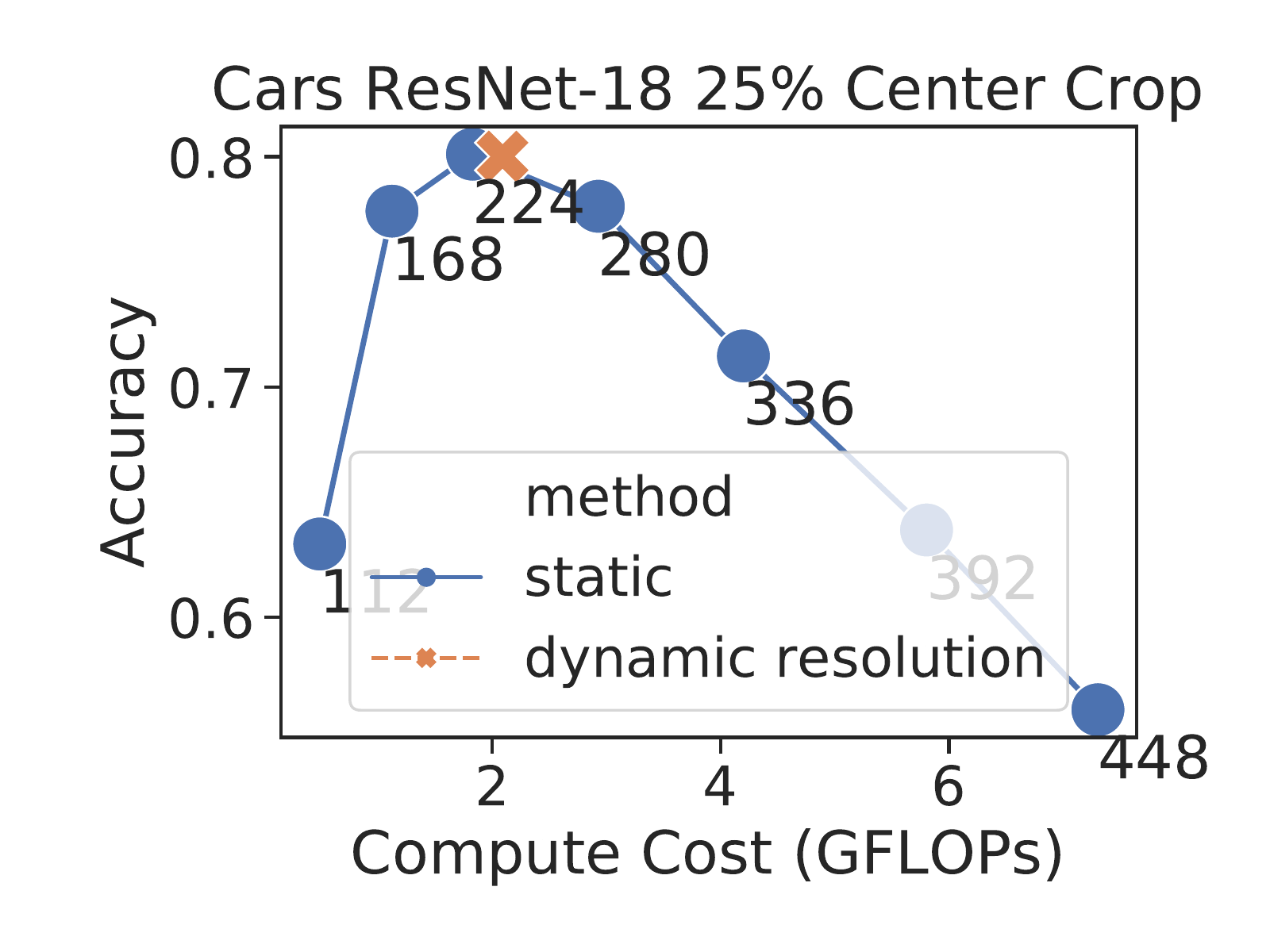} \\
    \small (a)
    \end{tabular}
    \begin{tabular}{@{}c@{}}
    \includegraphics[width=0.245\textwidth]{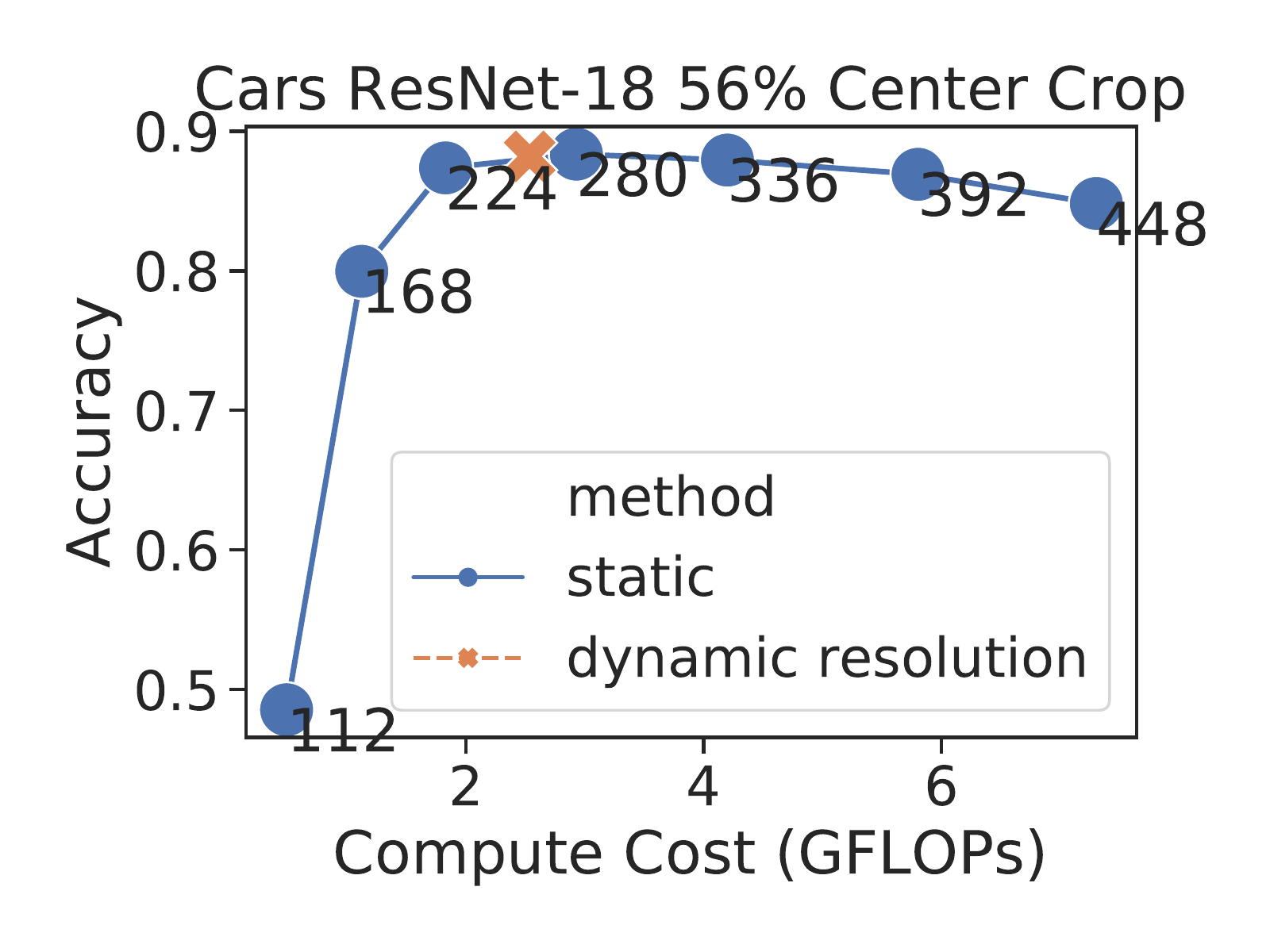} \\
    \small (b)
    \end{tabular}
    \begin{tabular}{@{}c@{}}
    \includegraphics[width=0.245\textwidth]{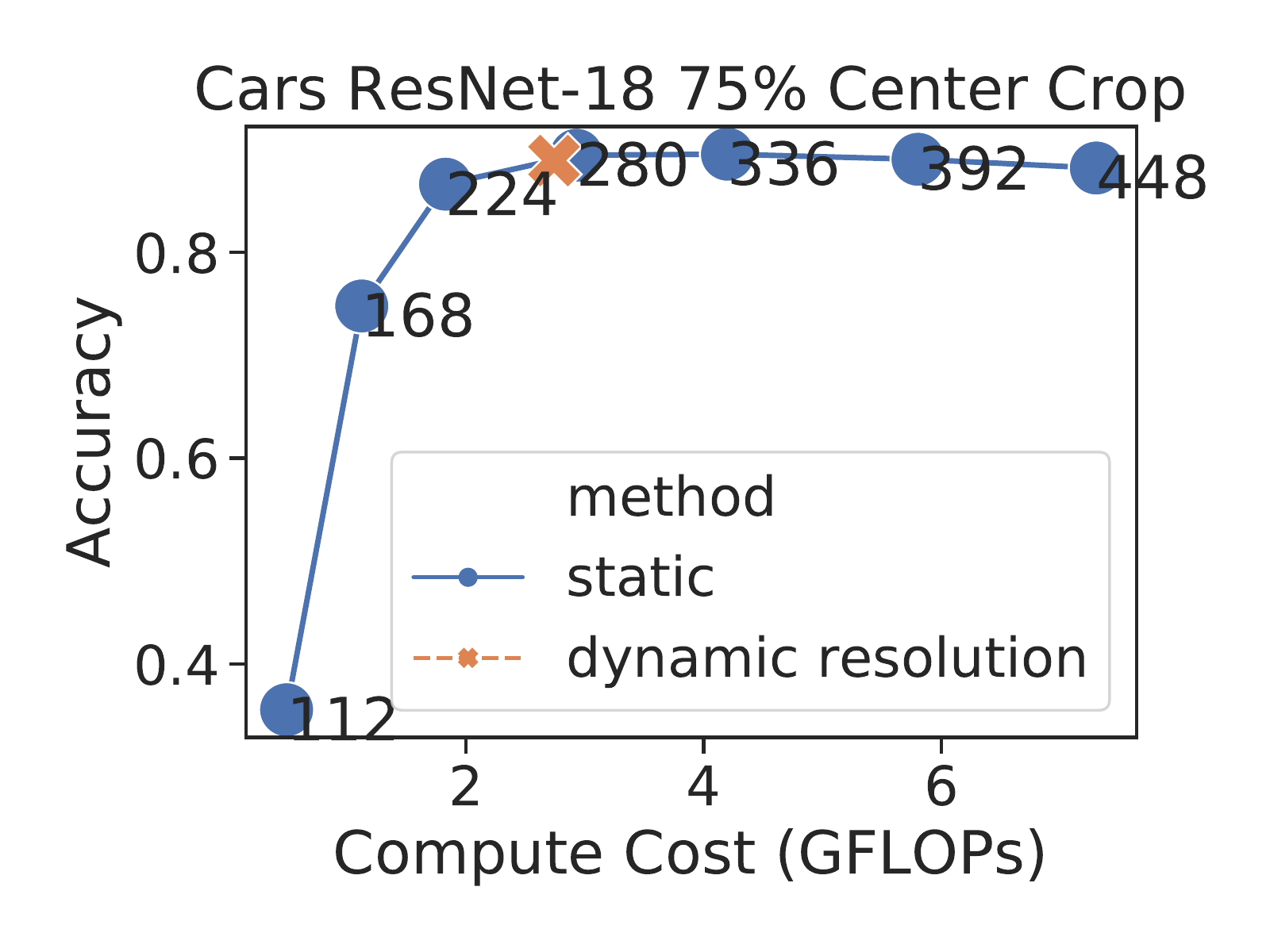} \\
    \small (c)
    \end{tabular}
    \begin{tabular}{@{}c@{}}
    \includegraphics[width=0.245\textwidth]{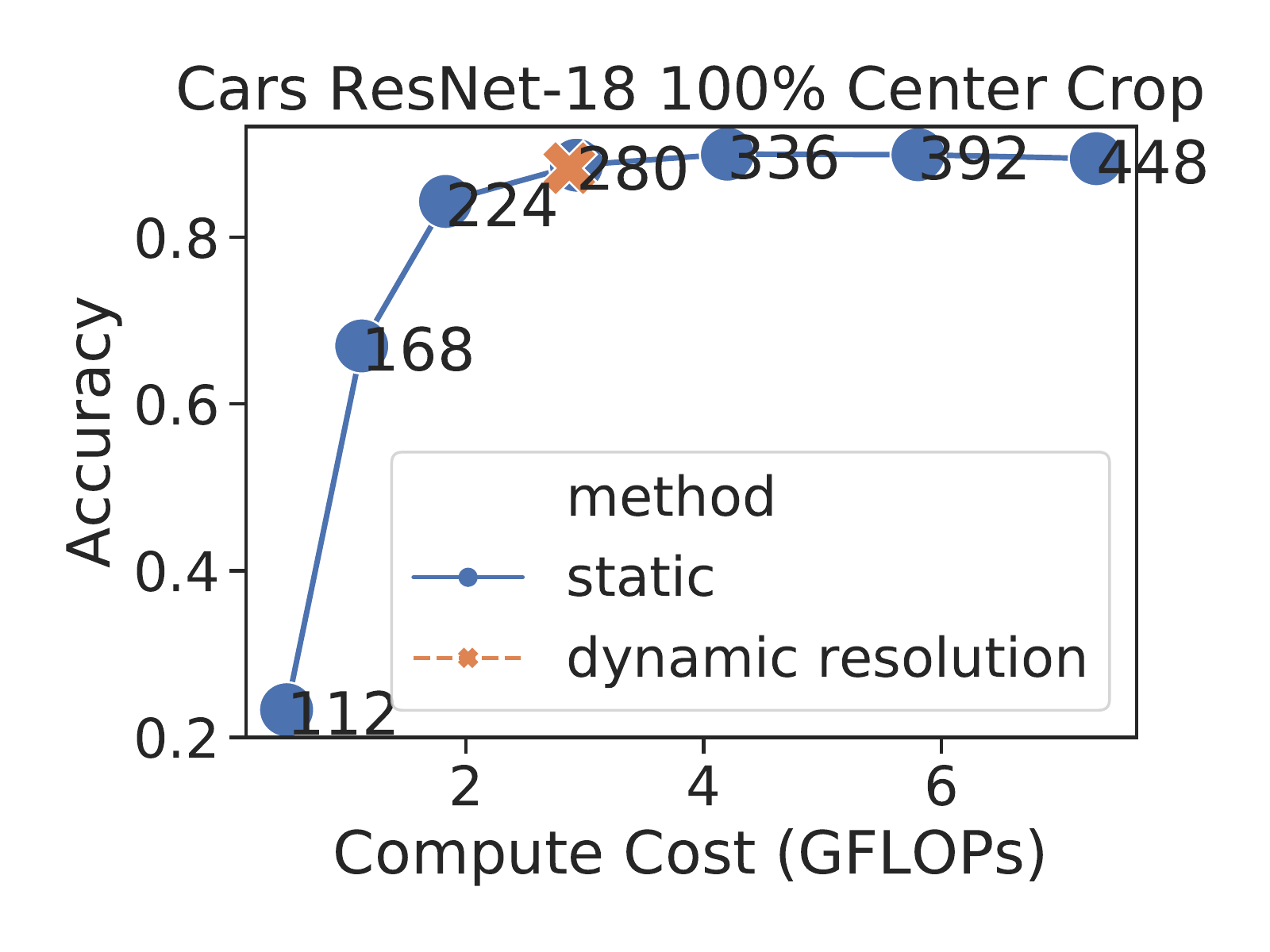} \\
    \small (d)
    \end{tabular}
    \begin{tabular}{@{}c@{}}
    \includegraphics[width=0.245\textwidth]{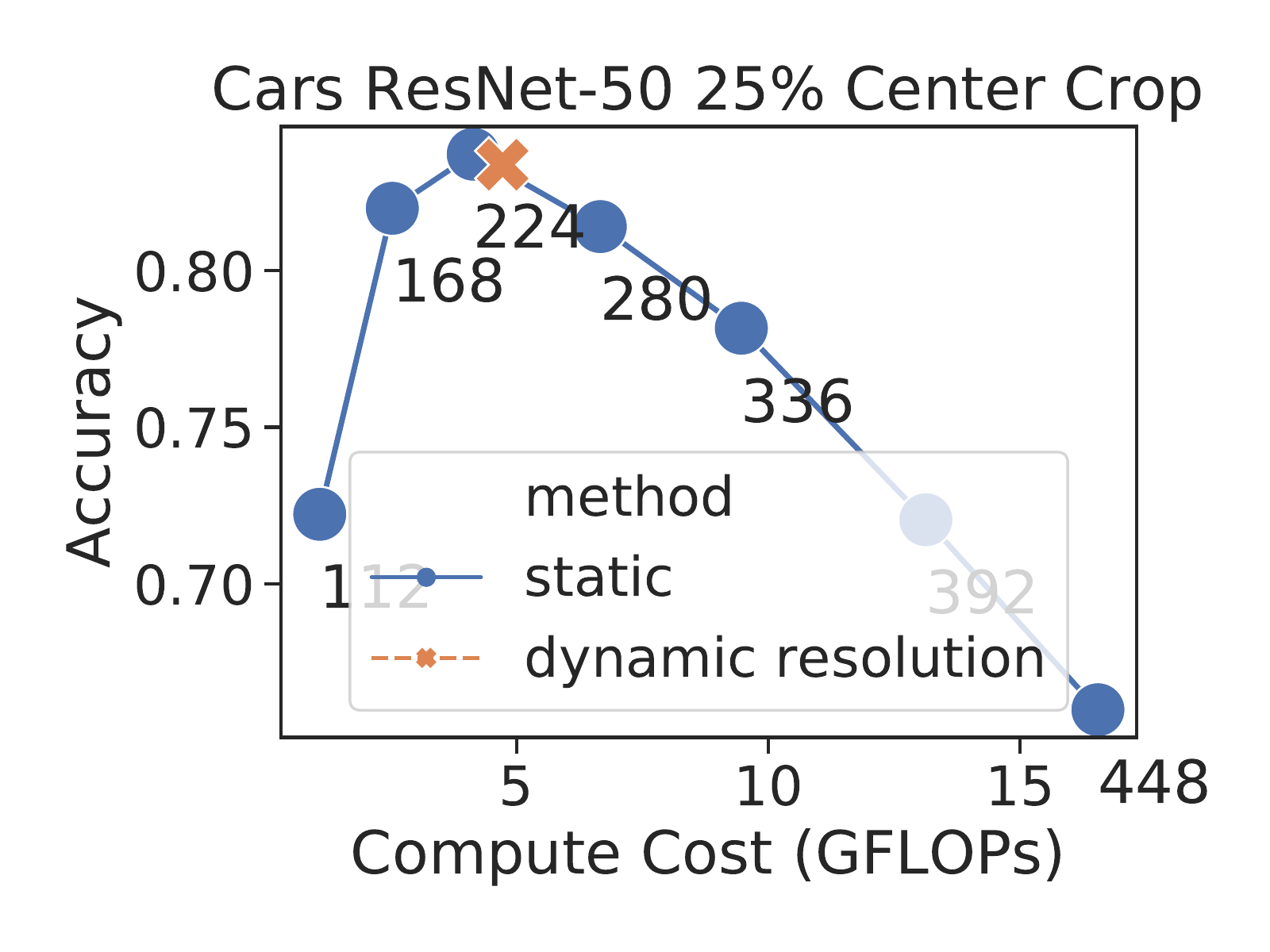} \\
    \small (e)
    \end{tabular}
    \begin{tabular}{@{}c@{}}
    \includegraphics[width=0.24\textwidth]{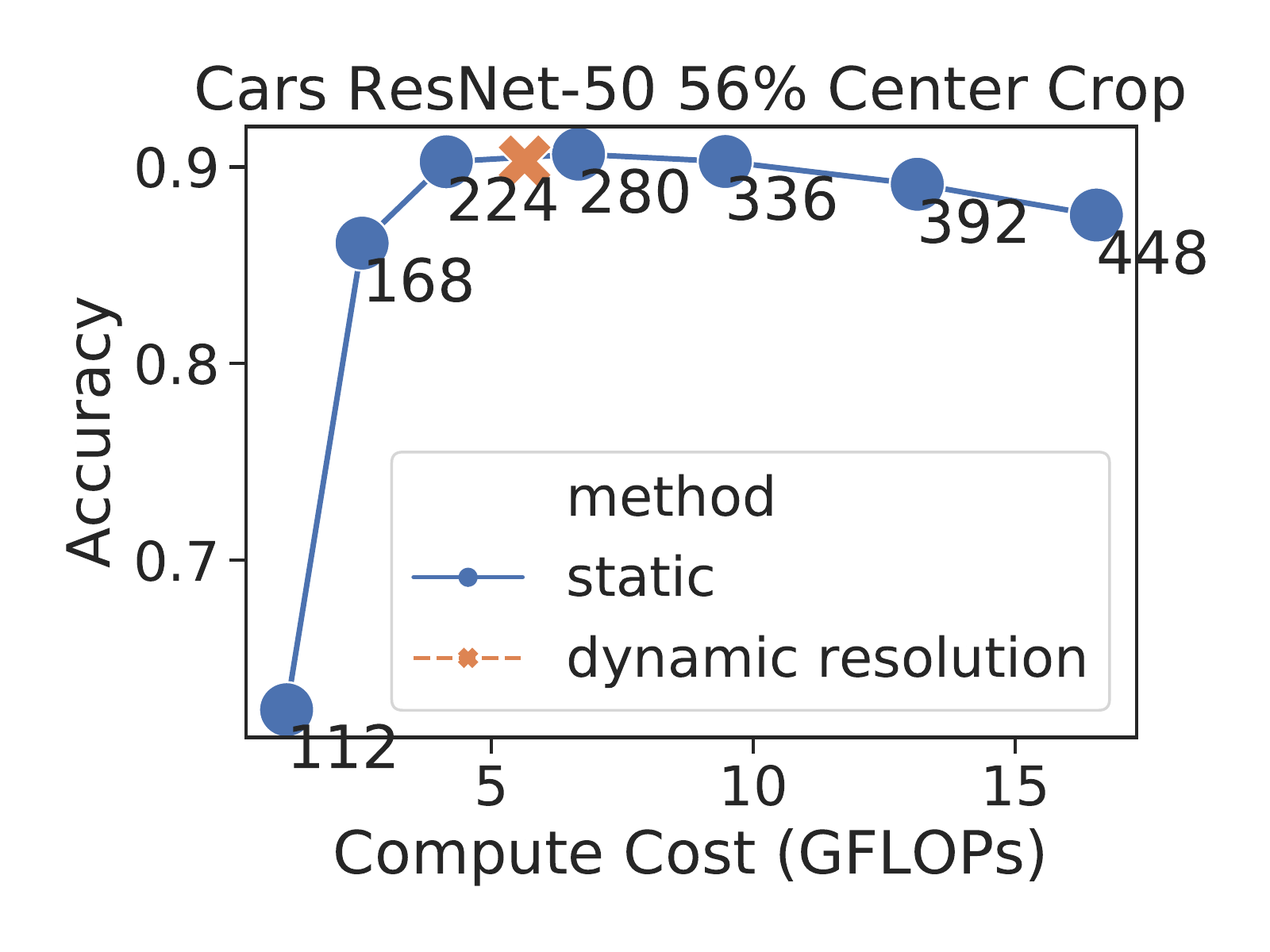} \\
    \small (f)
    \end{tabular}
    \begin{tabular}{@{}c@{}}
    \includegraphics[width=0.24\textwidth]{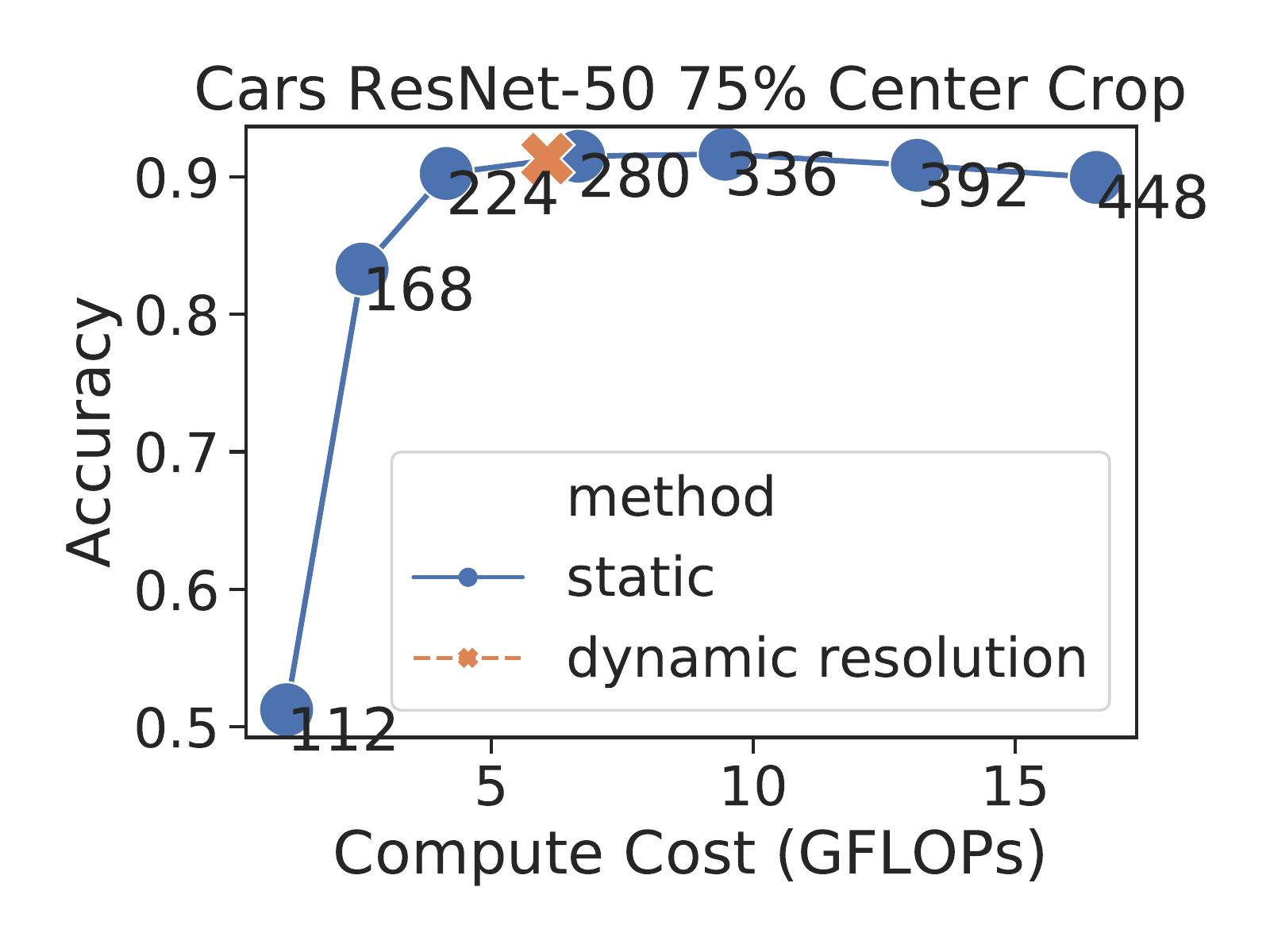} \\
    \small (g)
    \end{tabular}
    \begin{tabular}{@{}c@{}}
    \includegraphics[width=0.24\textwidth]{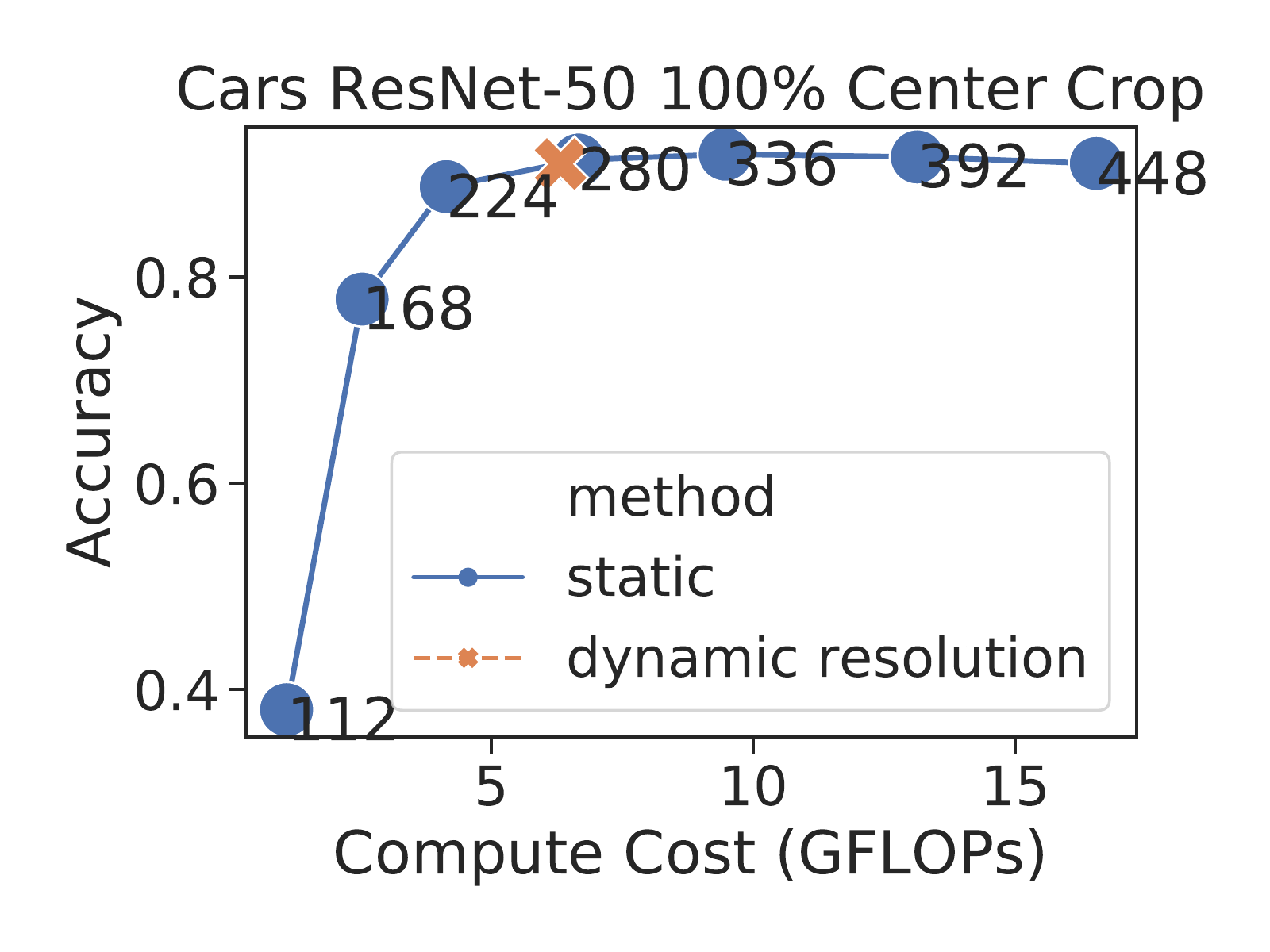} \\
    \small (h)
    \end{tabular}
    \caption{Accuracy vs. average FLOPs with static and dynamic resolution using ResNet-18 (a-d)/50 (e-h) on Cars. Crop sizes increase left to right from 25-100\%. Smaller crops favor lower resolutions while larger crops favor higher resolutions due to the models' dependence on object scale. The dynamic resolution approach operates near the apex of each curve.}
    \label{fig:accflops_resnet_cars}
\end{figure*}

\begin{table*}[!ht]
    \normalsize
    \centering
    \begin{tabular}{c|}
    ImageNet\\
    Res \\
    \hline
    $112$ \\ 
    $168$ \\ 
    $224$ \\ 
    $280$ \\ 
    $336$ \\ 
    $392$ \\ 
    $448$ \\
    dynamic\\
    \end{tabular}
    \begin{tabular}{|c|c|}
    \multicolumn{2}{|c|}{ ResNet-18 75\% Crop}\\
    Default & Calibrated \\
    \hline
    47.8 & 47.7\\
    62.7 & 62.7\\ 
    69.5 & 69.5\\ 
    70.7 & 70.7\\ 
    70.1 & 70.1\\ 
    69.4 & 69.4\\ 
    68.9 & 69.0\\
    70.6 & 70.6\\ 
    \end{tabular}
    \begin{tabular}{|c|c|}
    \multicolumn{2}{|c|}{ ResNet-18 56\% Crop}\\
    Default & Calibrated  \\
    \hline
    49.9 & 49.8 \\ 
    62.9 & 62.8 \\ 
    68.7 & 68.7 \\ 
    69.6 & 69.5 \\ 
    68.6 & 68.5 \\ 
    67.4 & {\color{red}67.2} \\ 
    66.6 & 66.5 \\
    69.6 & {\color{red}69.4} \\ 
    \end{tabular}
    \begin{tabular}{|c|c|}
    \multicolumn{2}{|c|}{ ResNet-18 25\% Crop}\\
    Default & Calibrated  \\
    \hline
    49.4 & 49.3 \\ 
    57.7 & 57.7 \\ 
    61.4 & 61.4 \\ 
    60.9 & {\color{red} 60.7} \\ 
    58.2 & {\color{red} 57.5} \\ 
    55.3 & {\color{red} 54.7} \\ 
    52.9 & {\color{red} 52.6} \\
    61.6 & 61.5 \\ 
    \end{tabular}
    \begin{tabular}{|c}
    \\
    Read Savings \\
    \hline
    16.4\%\\ 
    5.1\%\\ 
    5.8\%\\ 
    20.2\%\\ 
    27.7\%\\ 
    17.8\%\\ 
    9.5\% \\
    11.2,10.6,8.9\%\\ 
    \end{tabular}

    \begin{tabular}{c|}
    ImageNet\\
    Res \\
    \hline
    $112$ \\ 
    $168$ \\ 
    $224$ \\ 
    $280$ \\ 
    $336$ \\ 
    $392$ \\ 
    $448$ \\
    dynamic\\
    \end{tabular}
    \begin{tabular}{|c|c|}
    \multicolumn{2}{|c|}{ ResNet-50 75\% Crop}\\
    Default & Calibrated \\
    \hline
    58.2 & 58.1\\
    70.5 & 70.5\\ 
    74.9 & 74.9\\ 
    76.0 & 76.0\\ 
    75.3 & 75.3\\ 
    74.7 & 74.7\\ 
    74.2 & 74.2\\
    75.7 & 75.6\\ 
    \end{tabular}
    \begin{tabular}{|c|c|}
    \multicolumn{2}{|c|}{ ResNet-50 56\% Crop}\\
    Default & Calibrated  \\
    \hline
    60.0 & 60.0\\ 
    70.5 & 70.5\\ 
    73.9 & 73.9\\ 
    74.5 & 74.6\\ 
    74.0 & 74.0\\ 
    73.2 & 73.1\\ 
    72.4 & 72.3\\
    74.3 & 74.3 \\ 
    \end{tabular}
    \begin{tabular}{|c|c|}
    \multicolumn{2}{|c|}{ ResNet-50 25\% Crop}\\
    Default & Calibrated  \\
    \hline
    58.5 & 58.5 \\ 
    65.4 & 65.4\\ 
    67.6 & 67.5\\ 
    67.1 & 67.0 \\ 
    65.8 & 65.7\\ 
    63.5 & 63.2 \\ 
    60.7 & {\color{red} 60.4}\\
    67.5 & 67.5 \\ 
    \end{tabular}
    \begin{tabular}{|c}
    \\
    Read Savings \\
    \hline
    7.4\%\\ 
    2.1\%\\ 
    8.9\%\\ 
    19.2\%\\ 
    6.2\%\\ 
    9.1\%\\ 
    8.0\% \\
    6.8,6.7,6.5\%\\ 
    \end{tabular}
    \caption{ImageNet read bandwidth savings: tables compare accuracy when reading all data vs. reading the quantity of data according to storage calibration. Accuracy degradation $> 0.1\%$ highlighted. Read savings for the dynamic pipeline are for each crop size.
    Note that the read savings are identical for each crop size as we do not store pre-cropped versions of each image, opting instead to read a different number of scans for each image based on calibrated quality thresholds for each resolution.
    }
    \label{tab:imagenet_storage}
\end{table*}

\begin{table*}[!ht]
    \normalsize
    \centering
    \begin{tabular}{c|}
    Cars \\
    Res \\
    \hline
    $112$ \\ 
    $168$ \\ 
    $224$ \\ 
    $280$ \\ 
    $336$ \\ 
    $392$ \\ 
    $448$ \\
    dynamic\\
    \end{tabular}
    \begin{tabular}{|c|c|}
    \multicolumn{2}{|c|}{ ResNet-18 75\% Crop}\\
    Default & Calibrated \\
    \hline
    35.6 & 35.6 \\
    74.8 & 74.7\\ 
    86.6 & 86.6\\ 
    89.4 & 89.4\\ 
    89.5 & 89.5\\ 
    89.0 & 89.0\\ 
    88.2 & 88.1\\
    88.9 & 88.9\\ 
    \end{tabular}
    \begin{tabular}{|c|c|}
    \multicolumn{2}{|c|}{ ResNet-18 56\% Crop}\\
    Default & Calibrated  \\
    \hline
    48.6 & 48.6 \\ 
    80.0 & {\color{red}79.6}\\ 
    87.4 & 87.3 \\ 
    88.4 & 88.4 \\ 
    87.9 & 88.0 \\ 
    86.9 & 86.9 \\ 
    84.8 & 84.7\\
    88.2& 88.2 \\ 
    \end{tabular}
    \begin{tabular}{|c|c|}
    \multicolumn{2}{|c|}{ ResNet-18 25\% Crop}\\
    Default & Calibrated  \\
    \hline
    63.2 & 63.1 \\ 
    77.6 & {\color{red}77.3}\\ 
    80.1 & 80.1\\ 
    77.9 & 77.9 \\ 
    71.3 & 71.4 \\ 
    63.8 & 63.8 \\ 
    56.0 & {\color{red}55.8} \\
    80.0 & 80.0 \\ 
    \end{tabular}
    \begin{tabular}{|c}
    \\
    Read Savings \\
    \hline
    31.8\%\\ 
    59.4\%\\ 
    20.5\%\\ 
    29.8\%\\ 
    31.4\%\\ 
    37.2\%\\ 
    43.0\% \\
    25.2,24.0,21.6\%\\ 
    \end{tabular}
    
    \begin{tabular}{c|}
    Cars\\
    Res \\
    \hline
    $112$ \\ 
    $168$ \\ 
    $224$ \\ 
    $280$ \\ 
    $336$ \\ 
    $392$ \\ 
    $448$ \\
    dynamic\\
    \end{tabular}
    \begin{tabular}{|c|c|}
    \multicolumn{2}{|c|}{ ResNet-50 75\% Crop}\\
    Default & Calibrated \\
    \hline
    51.2 & {\color{red} 50.8} \\
    83.3 & 83.3\\ 
    90.2 & 90.2\\ 
    91.5 & 91.4\\ 
    91.6 & 91.6\\ 
    90.8 & 90.8\\ 
    90.0 & 89.9\\
    91.3 & 91.2\\ 
    \end{tabular}
    \begin{tabular}{|c|c|}
    \multicolumn{2}{|c|}{ ResNet-50 56\% Crop}\\
    Default & Calibrated  \\
    \hline
    62.4 & {\color{red}62.0} \\ 
    86.1 & 86.1 \\ 
    90.3 & 90.2 \\ 
    90.6 & 90.6 \\ 
    90.3 & 90.3 \\ 
    89.1 & 89.1 \\ 
    87.6 & 87.5 \\
    90.3 & 90.2 \\ 
    \end{tabular}
    \begin{tabular}{|c|c|}
    \multicolumn{2}{|c|}{ ResNet-50 25\% Crop}\\
    Default & Calibrated  \\
    \hline
    72.2 & {\color{red} 71.5} \\ 
    82.0 & 81.9 \\ 
    83.7 & 83.6 \\ 
    81.4 & 81.4 \\ 
    78.2 & 78.1 \\ 
    72.0 & 71.9 \\ 
    66.0 & {\color{red}65.6} \\
    83.4 & 83.3 \\ 
    \end{tabular}
    \begin{tabular}{|c}
    \\
    Read Savings \\
    \hline
    68.8\%\\ 
    30.7\%\\ 
    40.9\%\\ 
    51.9\%\\ 
    6.5\%\\ 
    39.8\%\\ 
    49.3\% \\
    48.8,47.1,43.1\%\\ 
    \end{tabular}
    \caption{Cars read bandwidth savings; comparing accuracy when reading all data vs. reading the quantity of data according to storage calibration. Accuracy degradation $> 0.1\%$ highlighted.
    Read savings for the dynamic pipeline are for each crop size.
    Again, the read savings are identical for each crop size as we do not store pre-cropped versions of each image, opting instead to read a different number of scans for each image based on calibrated quality thresholds for each resolution.
    }
    \label{tab:cars_storage}
\end{table*}

\paragraph{Accuracy vs. FLOPs}
To highlight the flexibility of dynamic resolution compared to static approaches that perform inference at a fixed resolution, we compare the accuracy of dynamic resolution at several different center crop ratios (25\%, 56\%, 75\%, and 100\%).\footnote{``75\%'' corresponds to the common practice of selecting a center crop (e.g., of 224 pixels from a $256\times256$ image, or 448 pixels from a $512\times512$ image), though the true area is closer to 77\%.}
We give the baselines the advantage of knowing the distribution of object sizes in advance so the optimal crop for the whole dataset can be determined in advance, which may not be possible in real-world settings. 
The scale model in our two-model pipeline uses a MobileNet-v2~\cite{sandler2018mobilenetv2} architecture and corresponds to ~0.08 GFLOPs at a resolution of $112\times112$ compared to the 1.8 GFLOPs of ResNet-18 and 4.1 GFLOPs of ResNet-50 at $224\times224$, incurring only a small fraction of latency overhead compared to the backbone model. 

\autoref{fig:accflops_resnet_imagenet} and \autoref{fig:accflops_resnet_cars} show the accuracy achieved by static and dynamic resolution approaches across a range of crop sizes on ImageNet and Cars.
For ImageNet, the best static resolution for the 56\%, 75\%, and 100\% center crops was $280\times280$, as expected due to the use of random cropping during training favors slightly larger object scales.
Additionally, using the full crop (including \emph{more} of the image area) decreases model accuracy as object scales are biased towards smaller images.
On the other hand, the dynamic resolution pipeline attains most of the accuracy of the best static resolution for each approach at a lower FLOP cost, and is pareto-optimal and near the apex of accuracy for most resolution configurations.
We point out the large accuracy improvement achieved at the lowest resolution of ResNet-50 on Stanford Cars when switching from a 75\% to a 25\% crop; top-1 accuracy improves from roughly 50\% to above 70\% for the smaller center crop.
The accuracy drop with small crops for higher resolutions is much more dramatic in Cars than on ImageNet, as at a 25\% center crop, the accuracy at $448\times448$ is \emph{lower} than at $112\times112$ for Cars, but it remains higher for ImageNet. 

We find that from the perspective of robustness and computational cost, a dynamic resolution pipeline is a feasible and robust alternative to fine-tuning for a known distribution of object scales as previously proposed~\cite{touvron2019fixing}.
The use of a dynamic resolution approach can adapt to the choice of crop size, reducing the potential impact when the cropping or scale of images is unknown at inference time.

We use SSIM for storage calibration in our approach to find the optimal image quality to use for each resolution and crop size. We compare model accuracy and the amount of data read at several different crop sizes in \autoref{tab:imagenet_storage} and \autoref{tab:cars_storage}, against a baseline approach that reads the entirety of image data.
Overall, we find that calibration generalizes well, as we only observe slight accuracy losses when we lower image quality at smaller crop sizes across datasets. However, as expected, data savings can vary widely across datasets.
For example, only a few inference resolutions on ImageNet reach above 20\% data savings, whereas many resolution/model configurations on Stanford Cars reach 40\% savings with less than 0.1\% accuracy loss. With dynamic resolution, we achieve virtually no accuracy loss while saving up to 11.2\% and 48.8\% of read data on the ImageNet and Stanford Cars datasets respectively.  

Due to the use of the scale model, the potential data read savings of the dynamic resolution approach are bounded by the amount of data used at $112\times112$ (the resolution used by the scale model). To break this limit, it is possible to calibrate image quality for the scale model as well, using the same mechanism for the backbone model, as the scale model likely requires less image detail. We leave this as future work. 



\paragraph{Runtime Overhead of the Scale Model}
We benchmarked our untuned (no use of MKLDNN/autotuning) PyTorch scale model implementation at 9.7ms on 4790K corresponding to 30\% slowdown compared to tuned, static resolution ResNet-50 inference at $224\times224$, representing the worst case scenario as (1) autotuning can further reduce this overhead; (2) scale model overhead can be hidden as we can pipeline running scale model inference of the next batch with the backbone model inference of the current batch.

\section{Discussion}
Our evaluation highlighted the value of optimizing for model accuracy, storage bandwidth, and compute cost in tandem, as well as using a dynamic resolution during inference.
In this section, we extend our discussion to identify applications for our approach as well as alternatives.

\paragraph{Dynamic Resolution}
In scenarios where the distribution of object scales at test time is well known, using a static model and the appropriate center crop size is likely to yield a good trade between accuracy and computational cost.
Besides the apparent usefulness when used in situation where the object scale distribution is unknown during inference, we see the dynamic resolution approach as being useful when some form of load balancing or latency adjustment is desirable: one can adjust the crop size for inference to reduce the average computational cost of the model pipeline (e.g., a burst of requests causing long queuing delays) without affecting latency in general, as the scale model automatically compensates for the change in object scale. Additionally, the scale model also has the advantage of improving the robustness of the pipeline to the distribution of object scales.
\paragraph{Reducing Image Read Bandwidth Requirements}
In scenarios where costs are dependent on network and storage bandwidth (e.g., hosting inference applications in the cloud), it is beneficial to reduce the amount of data transferred.
Using either a static resolution approach (or a lightweight scale model on edge devices), a lower quality version of an image can be sent to save bandwidth costs.
Although we have also made the implicit assumption that no image data is discarded from storage, further savings can be obtained by (1) cropping, and (2) resizing images ahead of time if an inference system is chosen such that the storage--accuracy and FLOPs--accuracy tradeoff is to be made ahead of time.
Thus, we consider the bandwidth savings discussed in the evaluation section to be a lower bound in the absence of further domain information.

\paragraph{Quality Metrics}
Structural similarity is a crude proxy for image quality, especially for neural networks that favor quality metrics more in line with human perception~\cite{zhang2018unreasonable}.
Perhaps due to the choice of quality metric, higher resolutions potentially require \emph{lower} image quality or less image data than lower resolutions, further complicating the tradeoffs that can be made between accuracy, compute costs, and storage costs.
In this work, we have chosen aggressive targets for accuracy, aiming to reducing image data reads only when less than 0.05\% accuracy is lost.
More sophisticated quality metrics that map better to neural network perceptual quality, as well as reference free metrics~\cite{wang2011reduced} can further improve bandwidth savings or the computational efficiency of this approach.

\paragraph{Alternative Loss Functions for the Scale Model}
We note that the dynamic resolution model chooses resolutions solely based on their predicted accuracy given an input image.
It is possible to also incorporate inference costs (e.g., latency) for each resolution for further fine-tuning.
Still, using a two-model approach remains pareto-optimal in terms of accuracy vs. FLOPs while improving model accuracy over the default static resolution in most scenarios.



\section{Conclusion}
Image resolution is a fundamental hyperparameter in computer vision with ties to compute complexity, operator efficiency, and storage bandwidth requirements.
To enable the choice of resolution, we must account for the effects of other choices such as crop size and kernel implementation.
We systematically characterized the relationships between these choices, and describe methods for maximizing efficiency with respect to compute and storage cost.
We showed that up to 20-30 \% of image data can be omitted when reading from storage or transferring from the network without sacrificing accuracy and that resolution-specialized kernels allow for $\ge 1\%$ accuracy improvement at $1.2\times$--$1.7\times$ speedup.
Finally, we establish that dynamic resolution approach is a viable and efficient alternative to fine-tuning for a specific resolution and can compensate for variations in crop size that can appear at test-time without dramatically increasing inference time.
Overall, our message is that from a complex landscape of model hyperparameters and their tradeoffs, a rich set of optimization opportunities emerges.

{\small
\bibliographystyle{ieee_fullname}
\bibliography{main}
}

\end{document}